\RequirePackage[2023-07-04]{latexrelease}
\documentclass[lineno]{biometrika}

\RequirePackage[thmmarks]{ntheorem}
\makeatletter
\renewtheoremstyle{plain}
{\item[\hskip\labelsep \theorem@headerfont ##1\ \textup{##2}\theorem@separator]}
{\item[\hskip\labelsep \theorem@headerfont ##1\ \textup{##2}\ (##3)\theorem@separator]}
\makeatother
\let\latexdocument\document
\let\latexarabic\arabic
\let\document\latexdocument
\let\arabic\latexarabic
\def\rm{}
\nolinenumbers
\usepackage{amsmath}

\RequirePackage{amsfonts,amssymb}
\usepackage{float}
\usepackage{latexsym}
\usepackage{graphicx}
\usepackage{mathrsfs}
\usepackage{enumerate}
\usepackage{mathabx}
\usepackage{xr}
\usepackage{multirow}

\usepackage{times}
\usepackage{bm}
\usepackage{natbib}
\usepackage{caption}

\graphicspath{{./Figures/}}

\usepackage[plain,noend]{algorithm2e}

\newcommand{\bel}{\begin{eqnarray}\label}
\newcommand{\eel}{\end{eqnarray}}
\newcommand{\bes}{\begin{eqnarray*}}
\newcommand{\ees}{\end{eqnarray*}}
\newcommand{\bei}{\begin{itemize}}
\newcommand{\beiftnt}{\begin{itemize}\footnotesize}
\newcommand{\eei}{\end{itemize}}

\def\benu{\begin{enumerate}}
\def\eenu{\end{enumerate}}

\def\argmin{\mathop{\rm arg\, min}}
\def\real{{\mathbb{R}}}

\def\E{{\mathbb{E}}}
\def\P{{\mathbb{P}}}

\def\complex{\mathop{{\rm I}\kern-.58em\hbox{\rm C}}\nolimits}

\def\rank{\hbox{\rm rank}}

\def\mathbold{\boldsymbol} 

\def\ba{\mathbold{a}}

\def\bb{\mathbold{b}}
\def\hbb{{\widehat{\bb}}}
\def\bB{\mathbold{B}}

\def\bC{\mathbold{C}}

\def\bE{\mathbold{E}}

\def\bi{\mathbold{i}}

\def\bI{\mathbold{I}}

\def\bM{\mathbold{M}}

\def\bu{\mathbold{u}}
\def\hbu{{\widehat{\bu}}}
\def\bU{\mathbold{U}}\def\hbU{{\widehat{\bU}}}

\def\bv{\mathbold{v}}
\def\hbv{{\widehat{\bv}}}
\def\bV{\mathbold{V}}\def\hbV{{\widehat{\bV}}}

\def\bX{\mathbold{X}}\def\hbX{{\widehat{\bX}}}\def\tbX{{\widetilde{\bX}}}

\def\by{\mathbold{y}}

\def\bZ{\mathbold{Z}}\def\hbZ{{\widehat{\bZ}}}

\def\ep{\varepsilon}\def\eps{\epsilon}
\def\bep{\mathbold{\ep}}

\def\bTheta{\mathbold{\Theta}}

\def\lam{\lambda}

\def\hlambda{\widehat{\lambda}}

\def\bSigma{\mathbold{\Sigma}}\def\hbSigma{{\widehat{\bSigma}}}

\def\bOmega{\mathbold{\Omega}}

\def\T{{ \mathrm{\scriptscriptstyle T} }}

\def\bmX{{\bf{\mathcal{X}}}}

\def\bmB{\mathbf{\mathcal{B}}}
\def\bmA{\mathbf{\mathcal{A}}}
\def\bmC{\mathbf{\mathcal{C}}}

\def\argmin{\operatorname{argmin} \displaylimits}

\def\mR{\mathcal{R}}

\def\mT{\mathcal{T}}

\def\E{\mathbb{E}}

\def\T{\top}

\def\tr{\text{tr}}

\def\tvec{\text{vec}}

\def\hmC{\widehat{\bf{\mathcal{C}}}}

\def\hbZ{\widehat{\bZ}}

\def\dist{\text{dist}}

\def\barX{\widebar{\bX}}
\def\barB{\widebar{\bB}}
\def\barb{\widebar{\bb}}

\def\argmin{\mathop{\rm arg\, min}}
\def\real{\mathop{{\rm I}\kern-.2em\hbox{\rm R}}\nolimits}

\usepackage{hyperref}

\usepackage{cleveref}
\crefname{section}{§}{§§}
\Crefname{section}{§}{§§}

\makeatletter
\renewcommand{\algocf@captiontext}[2]{\quad #1\algocf@typo. \AlCapFnt{}#2} 
\def\@algocf@capt@plain{top}
\renewcommand{\algocf@makecaption}[2]{%
	\addtolength{\hsize}{\algomargin}%
	\sbox\@tempboxa{\algocf@captiontext{#1}{#2}}%
	\ifdim\wd\@tempboxa >\hsize
	\hskip .5\algomargin%
	\parbox[t]{\hsize}{\algocf@captiontext{#1}{#2}}
	\else%
	\global\@minipagefalse%
	\hbox to\hsize{\box\@tempboxa}
	\fi%
	\addtolength{\hsize}{-\algomargin}%
}
\makeatother

\begin{document}

	\jname{Biometrika}
	\jyear{2024}
	\jvol{111}
	\jnum{2}
	
	
	\markboth{Feng and Yang}{Deep Kronecker Network}
	
	\title{Deep Kronecker Network}
	
	\author{Long Feng}\affil{Department of Statistics and Actuarial Science, The University of Hong Kong\\ Pokfulam Road, Hong Kong \email{lfeng@hku.hk}}
	
	\author{Guang Yang}\affil{School of Data Science, City University of Hong Kong\\ 83 Tat Chee Ave, Kowloon Tong, Hong Kong \email{guang.yang@my.cityu.edu.hk}}
	
	\maketitle

\begin{abstract}
We propose a novel framework called Deep Kronecker Network, designed for analyzing medical imaging data, such as MRI, fMRI, CT, etc. Medical imaging data differs from general images in at least two aspects: i) sample size is typically considerably smaller, ii) model interpretation is more of a concern compared to outcome prediction. As such, general methods are difficult to be applied directly. The proposed Deep Kronecker Network is built on a Kronecker product structure and implicitly imposes a piecewise smooth property on coefficients, which allows it to adapt to low sample size and provide desired model interpretation. This approach is general in the sense that it works for both matrix and tensor represented image data, and could be applied to both continuous and discrete outcomes. Moreover, the Kronecker structure can be written into a convolutional form, so Deep Kronecker Network resembles a CNN, particularly, a fully convolutional network. Interestingly, Deep Kronecker Network is also highly connected to the tensor regression framework proposed by Zhou et al. (2013), where a low-rank structure is imposed on tensor coefficients. We conduct both classification and regression analyses using real MRI data from the Alzheimer’s Disease Neuroimaging Initiative to demonstrate the effectiveness of Deep Kronecker Network.
\end{abstract}


\section{Introduction}
Medical imaging analysis plays a central role in modern medicine. The advancement of imaging technologies have tremendously benefited the diagnosis and treatment of diseases.

Although image analysis has been intensively studied over the past decades, medical image data is significantly different from general images in at least two aspects. First, the sample size is typically considerably smaller, while the image data are of higher order and higher dimension. In MRI analysis, for instance, it is common to encounter datasets comprising merely hundreds or at most thousands of patients, each having an MRI scan consisting of millions of voxels.
As a comparison, in general image recognition or computer vision problems, the sample size can easily reach millions, surpassing the image dimensions significantly.
Second, while many image recognition problems prioritize outcome prediction, medical imaging analysis places greater emphasis on model interpretability.

Due to the unique nature of medical imaging data, it is difficult to apply general image methods directly. 
CNN \citep[][]{fukushima1982neocognitron,lecun1998gradient} is arguably the most successful method for image recognition in recent years. However, its training requires large amount of samples, which is hardly available in medical imaging analysis. Additionally, a CNN model, with numerous unknown parameters presented in a ``black box'', is extremely difficult to interpret and cannot meet the requirements of medical imaging analysis.

Within the statistics community, numerous endeavors have been made to develop methodologies for medical imaging analysis. A common strategy involves vectorizing the images and utilizing the resulting pixels as independent predictors. Based on this strategy, various methods have been developed in the literature, such as Total Variation and fused Lasso based approaches \citep{rudin1992nonlinear, wang2017generalized,tibshirani2005sparsity}, Bayesian methods \citep{goldsmith2014smooth, kang2018scalar}, etc. In spite of their effectiveness in different applications, vectorizing the images is clearly not an optimal strategy. Not mentioning the loss of spatial information, the resulting ultra high-dimensional vectors also face severe computational limitations.
When image data are represented as tensors, \citet{zhou2013tensor} proposed a tensor regression framework that imposes a canonical polyadic (CP) low-rank structure on the tensor coefficients, with which the number of unknown parameters could be significantly reduced. Built on that, \citet{feng2021brain} further proposed a new Internal Variation penalization to mimic the effects of Total Variation and promote smoothness of image coefficients. While the tensor regression framework is appealing, it is designed for general tensor represented predictors, and does not fully utilize the special nature of image data. Recently, \citet{wu2022sparse} proposed an innovative framework named Sparse Kronecker Product Decomposition to detect signal regions in image regression. While this approach is specifically designed for sparse signal detection, it is not well-suited for the analysis of images with dense signals.

To this end, it is desired to develop an approach for medical imaging analysis that is able to i) adapt to low sample size limitation, ii) enjoy good interpretability, and iii) achieve desired prediction power.
In this paper, we develop a novel framework named Deep Kronecker Network (DKN) that is able to achieve all three goals. Deep Kronecker Network is built on a Kronecker product structure and implicitly imposes a latent piecewise smooth property of coefficients. This enables us to locate the image regions that are most influential to the outcome, facilitating model interpretation. Deep Kronecker Network works for both matrix and (high-order) tensor represented image data, so CT, MRI, fMRI and other types of medical imaging data could all be handled. Furthermore, Deep Kronecker Network is embedded in a generalized linear model, therefore it is applicable to both discrete and continuous responses. From these two points, Deep Kronecker Network is a general approach.

We call Deep Kronecker Network a network because it resembles a CNN, particularly, a fully convolutional network. While Deep Kronecker Network originates from a Kronecker structure, it could also be written into a convolutional form.
But different from classical CNN, the convolutions in Deep Kronecker Network have no overlaps. This design not only allows us to achieve maximized dimension reduction, but also provides desired model interpretability.
Interestingly, Deep Kronecker Network is also connected to the tensor regression framework of \citet{zhou2013tensor}. We show that Deep Kronecker Network not only includes Zhou's tensor regression as a special case, it could also be easily implemented by applying Zhou's tensor regression on reshaped images.
Therefore, the three seemingly irrelevant methods, fully convolutional network, tensor regression and Deep Kronecker Network could be connected.
Finally, we implemented a real MRI analysis from Alzheimer's Disease Neuroimaging Initiative to further demonstrate the effectiveness of Deep Kronecker Network.

\section{Deep Kronecker Network}\label{sec:dkn}
Suppose that we observe $n$ samples with tensor represented images $\bmX_i\in\mathbb{R}^{d\times p\times q}$ and scalar responses $y_i$, for any $i\in[n]$. Assume that $y_i$ follows a generalized linear model: 
\begin{equation}\label{model1}
	y_i|\bmX_i \sim \P(y_i|\bmX_i) = \rho (y_i)\exp\Big\{y_i\ \langle\bmX_i, \bmC\rangle-\psi\big(\langle\bmX_i, \bmC\rangle\big)\Big\},
\end{equation}
where 
$\bmC\in\mathbb{R}^{d\times p \times q}$ is the target unknown coefficient tensor, $\langle\cdot,\cdot\rangle$ is the inner product, $\rho(\cdot)$ and $\psi(\cdot)$ are certain known univariate functions. 
In model (\hyperref[model1]{1}), we focus on the image analysis and omit other potential design variables, such as age, sex, etc. They can be added back to the model easily if necessary. Given model (\hyperref[model1]{1}), we have that for a certain known link function $g(\cdot)$,
\begin{equation}\label{model0}
g\left\{\E(y_i)\right\}=\left\langle\bmX_i,\bmC\right\rangle.  
\end{equation}
To get started, we introduce the Kronecker product for $K$-order tensors.
\begin{definition}\label{def1}
	(Tensor Kronecker Product) Let $\bmA\in\mathbb{R}^{p_1\times \cdots\times p_K}$ and $\bmB\in\mathbb{R}^{q_1 \times \cdots \times q_K}$ be two $K$-order tensors with entries denoted by $\bmA_{i_1,\ldots,i_K}$ and $\bmB_{j_1,\ldots,j_K}$, respectively. Then the tensor Kronecker product $\bmC=\bmA \otimes \bmB$ is defined by $\bmC_{[j_1 i_1],\ldots, [j_K i_K]} = \bmA_{i_1,\ldots,i_K} \bmB_{j_1,\ldots,j_K}$ 
	for all possible values of $(i_1,\ldots,i_K)$ and $(j_1,\ldots,j_K)$, where $[j_k i_k]=j_k + (i_k-1)q_k$ for all $k \in [K]$.
\end{definition}

Under the framework of Deep Kronecker Network, we propose to model the coefficient tensor $\bmC$ with a rank-R Kronecker product decomposition with $L(\ge 2)$ factors:
\begin{equation}\label{kk1}
	\bmC=\sum_{r=1}^R\bmB^r_{L}\otimes\bmB^r_{L-1} \otimes \cdots\otimes \bmB^r_{1},
\end{equation}
where $\bmB^r_{l}\in\mathbb{R}^{d_l\times p_l \times q_l}$ are unknown tensors for all $l\in[L]$ and $r\in[R]$, and referred to Kronecker factors. The sizes of $\bmB^r_{l}$ are unknown, but are assumed to satisfy $d=\prod_{l=1}^L d_l$, $p=\prod_{l=1}^L p_l$ and $q=\prod_{l=1}^L q_l$. For ease of notation, we also write (\hyperref[kk1]{3}) into the form $\bmC= \sum_{r=1}^R \bigotimes_{l=L}^1 \bmB_l^r$.
 
\begin{figure}[t]
	\centering
	\includegraphics[width=0.5\textwidth]{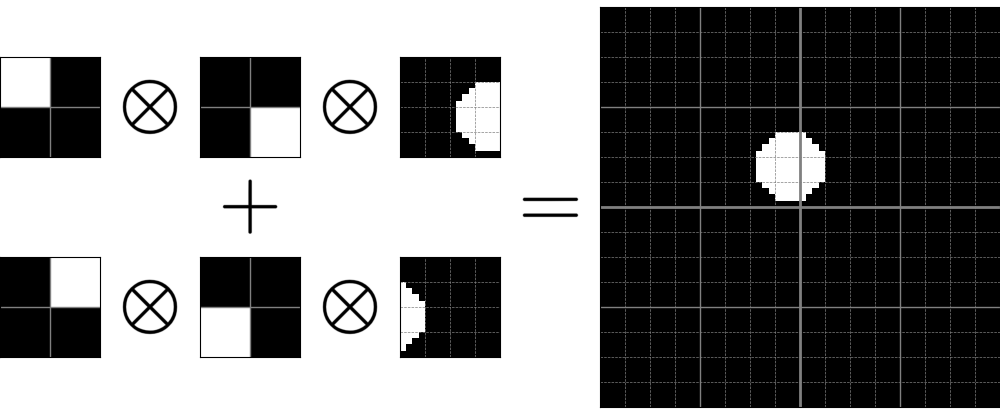}
	\caption{An illustration of DKN with $L=3$, $R=2$, $\bmB_3^r, \bmB_2^r\in\mathbb{R}^{2\times 2}$, $\bmB_1^r\in\mathbb{R}^{4\times 4}$, $r=1,2$.}
	\label{fig1}
	\vspace{-0.1in}
\end{figure}

Figure \hyperref[fig1]{1} illustrates a Deep Kronecker Network, suggesting a decomposition with a rank of $R=2$ and a factor number of $L=3$ for a sparse matrix wherein the signal takes the form of a circle. In general, (\hyperref[kk1]{3}) is able to approximate arbitrary matrices with a sufficiently large rank $R$. This can be seen by relating (\hyperref[kk1]{3}) to CP decomposition; see \cref{sec:tr}.

Deep Kronecker Network is designed for medical image analysis with low-sample-size and high-dimensional data.
It could reduce the parameter number from $\prod_{l=1}^L d_lp_lq_l$ to $R\sum_{l=1}^L d_lp_lq_l$. Considering that the sample sizes in many medical image analyses are only in the hundreds or thousands, such dimension reduction becomes more significant and critical.

Within the literature, Kronecker product decomposition has emerged as a powerful tool for matrix approximation and dimension reduction.
In particular, Kronecker product singular value decomposition is referred to the problem of recovering $\bB^r_{l}$ from a given matrix $\bC=\sum_{r=1}^R\bigotimes_{l=L}^1 \bB_l^r$, which was mostly studied when $L=2$, e.g., \citet{cai2019kopa}. 
While for general case with $L\ge 3$, it becomes a much more difficult problem \citep{hackbusch2005hierarchical}.  \citet{batselier2017constructive} considered its computation with $L\ge 3$ and proposed an algorithm to transform Kronecker product singular value decomposition to a CP decomposition problem. Besides, Kronecker product decomposition has also been studied in other contexts, e.g., correlation matrix estimation \citep{hafner2020estimation}, matrix autoregressive model \citep{chen2020constrained}, sparse signal detection \citep{wu2022sparse}, etc.

Given model (\hyperref[model1]{1}) to (\hyperref[kk1]{3}), 
we solve it with maximum likelihood estimation. 
For $y_i$ and $\bmX_i$, the negative likelihood function with regard to factors $\left[\bmB_1^{1},\ldots, \bmB_L^{R}\right]$ 
is proportional to
\begin{equation}\label{obj1}
	\ell\left(\bmB_1^{1},\ldots, \bmB_L^{R}\right) =\sum_{i=1}^n\left\{\psi\left(\left\langle\bmX_i, \sum_{r=1}^R\bigotimes_{l=L}^1 \bmB_l^r\right\rangle\right)-y_i\ \left\langle\bmX_i, \sum_{r=1}^R\bigotimes_{l=L}^1 \bmB_l^r\right\rangle\right\}.
\end{equation}
When the outcome $y_i$ is Gaussian distributed, the maximum likelihood reduces to ordinary least squares. Then optimization problem (\hyperref[obj1]{4}) could be solved by an alternating minimization algorithm to iteratively update the blocked factors $\left[\bmB^1_{l}, \bmB^2_{l},\ldots, \bmB^R_{l}\right]$, with $\left[\bmB^1_{l'}, \bmB^2_{l'},\ldots, \bmB^R_{l'}\right]$, $l'\neq l$ being fixed.
We defer the computation details to the Supplementary Material.

\section{DKN in convolutional form, FCN and nonlinear DKN}\label{sec:fcn}
To demonstrate the connection between Deep Kronecker Network and fully convolutional network, we first introduce a non-overlapping convolutional operator. For given tensors $\bmX\in \mathbb{R}^{d_0\times p_0\times q_0}$ and $\bmB\in \mathbb{R}^{d'\times p'\times q'}$, define the non-overlapping convolution between $\bmX$ and $\bmB$ as
\bes
\bmX*\bmB\in \mathbb{R}^{d''\times p''\times q''}, \ \ \ d''=d_0/d',\ \ p''=p_0/p', \ \ q''=q_0/q'
\ees
with the $(h,j,k)$-th component being
\bes
(\bmX*\bmB)_{h,j,k} = \left\langle\bmX_{h,j,k}^{d'\times p'\times q'}, \bmB \right\rangle, \ \ 1\le h\le d'', \ 1\le j\le p'', \ 1\le k\le q''.
\ees
Here $\bmX_{h,j,k}^{d'\times p'\times q'}$ is the $(h,j,k)$-th block of $\bmX$ and is of size $d'\times p'\times q'$. Then we have:
\begin{theorem}\label{fcneq}
	Deep Kronecker Network could be written into the convolutional form: 
	\bes
	g\left\{\E(y_i)\right\}=\left\langle\bmX_i, \sum_{r=1}^R\bigotimes_{l=L}^1 \bmB_l^r\right\rangle \ \ \ \Leftrightarrow \  \ \ g\left\{\E(y_i)\right\}= \sum_{r=1}^R\bmX_i * \bmB^r_{1}  * \bmB^r_{2} *\cdots * \bmB^r_{L-1} * \bmB^r_{L}.
	\ees
\end{theorem}

\begin{figure}[t]
	\centering
	\includegraphics[width=0.9\textwidth]{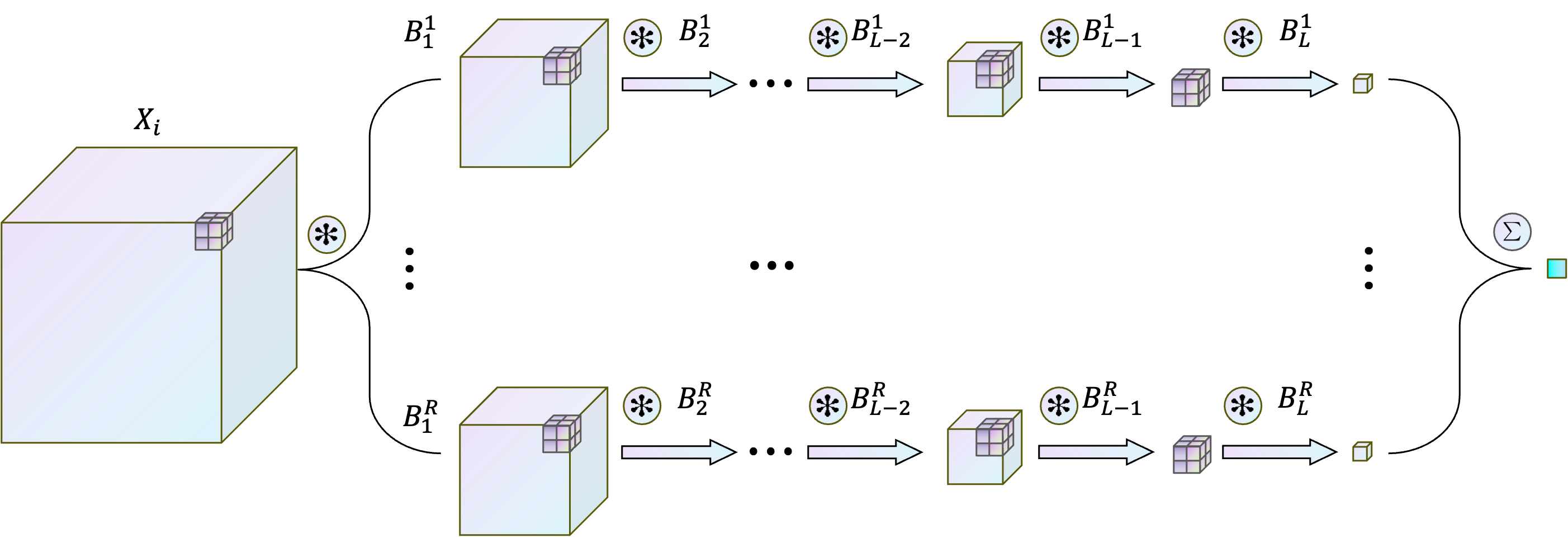}
	\caption{An illustration of DKN in a convolutional form.}
	\label{convolutionfigure}
	\vspace{-0.1in}
\end{figure}

Theorem \hyperref[fcneq]{1} implies the response $y_i$ is modeled by a summation of consecutive convolutions between image $\bmX_i$ and factors $\bmB^r_{l}$. In other words, Deep Kronecker Network could be viewed as a network with only convolutional layers.
More specifically, we may regard $L$ as the depth of a Deep Kronecker Network, $R$ as the width, and $\bmB_l^r$ as the convolution filters. 
But here the convolutions have no overlaps with each other, i.e., the stride sizes are equal to the filter sizes.
On one hand, the non-overlapping design makes Deep Kronecker Network to achieve maximized dimension reduction, thereby eliminating the need for pooling layers.
On the other hand, it allows for the explicit formulation of the coefficient tensor, enabling us to locate the significant regions and achieve desired model interpretability. Both aspects are important in medical imaging analysis. Figure. \hyperref[convolutionfigure]{2} illustrates Deep Kronecker Network in a convolutional form.

The activation function in Deep Kronecker Network is taken as an identity function. By introducing a nonlinear function, we can generalize it to its nonlinear version
\begin{equation}\label{nonlinear}
g\left\{\E(y_i)\right\}=\sum_{r=1}^Rh\left[ \cdots h\left\{h(\bmX_i * \bmB^r_{1})  * \bmB^r_{2}\right\} \cdots * \bmB^r_{L-1}\right] * \bmB^r_{L}.
\end{equation}
where $h(\cdot)$ is certain nonlinear activation function, e.g., ReLU.
The nonlinear Deep Kronecker Network could be solved easily using standard deep learning frameworks, such as {\it Pytorch}.

\section{DKN and Tensor Regression}\label{sec:tr}
	In this section, we demonstrate that Deep Kronecker Network not only includes tensor regression as a special case, it could also be easily implemented by applying tensor regression on reshaped images. 
	Suppose a three-order tensor $\bmC\in \mathbb{R}^{d\times p\times q}$ could be written as $\bmC=\bigotimes_{l=L}^1 \bmB_{l}$.
Then the entries of $\bmC$ are characterized by
	$\bmC_{{[h_1 \cdots h_L],[j_1 \cdots j_L],[k_1 \cdots k_L]}} = \prod_{l=1}^L [\bmB_l]_{h_l, j_l, k_l}$.
	The square brackets indicate grouping of indices. For example, the grouped index $[h_1\cdots h_L]$ is equivalent to the linear index $h_1+(h_2-1)d_1+\cdots+(h_L-1)\prod_{l=1}^{L}d_l$.
	
	Now let $\mT: \mathbb{R}^{d\times p\times q}\rightarrow \mathbb{R}^{(d_1p_1q_1)\times \cdots\times (d_Lp_Lq_L)}$ be a reshaping operator from tensor $\bmC$ to an $L$-order tensor $\mT(\bmC)$ with the entries characterized as below:
	\bes
	\left[\mT(\bmC)\right]_{{[h_1 j_1 k_1],\ldots,[h_L j_L k_L]}} =\bmC_{{[h_1 \cdots h_L],[j_1 \cdots j_L],[k_1 \cdots k_L]}}.
	\ees
	By this operator, \citet{batselier2017constructive} provides the following connection.
	\begin{lemma} \label{tkp}
		\citep{batselier2017constructive} Given a tensor $\bmC\in\mathbb{R}^{d\times p\times q}$, if
		$\bmC=\sum_{r=1}^R\bigotimes_{l=L}^1 \bmB_{l}^r$.
		then we have 
		$\mT(\bmC)=\sum_{r=1}^R\bb^r_{1}\circ\cdots\circ\bb^r_{L}$, where $\bb^r_{l}=\tvec(\bmB^r_l)$, for all $l\in[L]$ and $r\in[R]$.
	\end{lemma}

	As the reshaping operator $\mT(\cdot)$ is one-to-one and any tensor could be approximated by CP decomposition, Lemma \hyperref[tkp]{1} allows us to claim that Kronecker product decomposition (\hyperref[kk1]{3}) is also able to approximate arbitrary tensors.
	Built on Lemma \hyperref[tkp]{1}, we have the following theorem.
	\begin{theorem}\label{treq}
		The low-Kronecker-rank in Deep Kronecker Network is equivalent to a low-CP-rank assumption on the reshaped images $\mathcal{T}(\bmX_i)$. Let $\bb^r_{l}=\tvec\left(\bmB^r_{l}\right)$. Then we have
		\bes
		g\left\{\E(y_i)\right\}=\left\langle\bmX_i, \sum_{r=1}^R\bigotimes_{l=L}^1 \bmB_{l}^r\right\rangle \ \ \  \Leftrightarrow \ \ \  g\left\{\E(y_i)\right\}= \left\langle\mathcal{T}(\bmX_i), \sum_{r=1}^R\bb^r_{1}\circ\cdots\circ\bb^r_{L}\right\rangle.
		\ees 
	\end{theorem}
	\begin{figure}[t]
		\centering
		\includegraphics[width=0.9\textwidth]{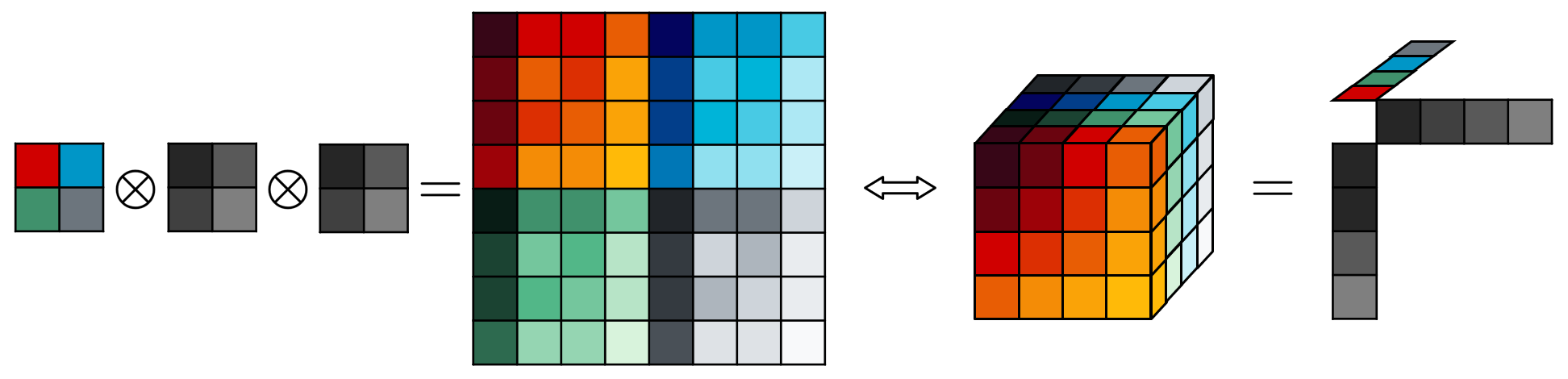}
		\caption{An illustration of connection between KPD and CPD.}
		\label{fig:trans}
		\vspace{-0.1in}
	\end{figure}
	\begin{remark} \label{treq2}
		Theorem \hyperref[treq]{2} suggests that Deep Kronecker Network could be solved by a two-step procedure: 1) reshape the original images, and 2) implement tensor regression, such as block relaxation algorithm in \cite{zhou2013tensor}, on the reshaped images. We note that the reshaping step is crucial to different performances of Deep Kronecker Network and tensor regression.
	\end{remark}

	\begin{remark} \label{treq3}
	Deep Kronecker Network includes tensor regression as a special case. Suppose that images are of size $D_1\times D_2\times D_3$. Then tensor regression could be viewed as a special Deep Kronecker Network with factor number $L=3$ and factors $\bmB_1^r\in\mathbb{R}^{D_1\times 1\times 1}$, $\bmB_2^r\in\mathbb{R}^{1\times D_2\times 1}$, $\bmB_3^r\in\mathbb{R}^{1\times 1\times D_3}$, for $r\in [R]$. Under such a case, $\mathcal{T}(\bmX_i)=\bmX_i$. Thus, Deep Kronecker Network is a more flexible and adaptive framework for allowing different sizes of factors.
	\end{remark}

	\begin{remark} \label{treq4}
	The size of factors $\bmB_{l}^r$ and number of layers $L_r$ are actually allowed to be different across $r$. In this situation, we could apply different reshaping operations $\mathcal{T}_r(\bmX_i)$ and obtain
	\bel{dkn2}
	g\left\{\E(y_i)\right\}
	=\sum_{r=1}^{R}\left\langle \mathcal{T}_r(\bmX_i), \bb_1^r \circ \cdots\circ \bb_{L_r}^r\right\rangle
	\eel
	Model (\hyperref[dkn2]{6}) is no longer in a form of tensor regression. But it still could be solved by alternating minimization algorithm with $\bb_l^r$ iteratively updated by fixing $\bb_{l'}^{r'}$, $l'\neq l$, $r'\neq r$.
	\end{remark}

	\begin{remark} \label{treq5}
	Deep Kronecker Network imposes a latent blockwise smoothness structure on the coefficients, which is particularly suitable for image data analysis.
	Figure \hyperref[fig:trans]{3} illustrates Kronecker product decomposition and its connection to CP decomposition. Evidently, the matrix produced by Kronecker product demonstrates a blockwise smooth (similar color) pattern.
	\end{remark}

	\section{Theoretical Analysis}\label{sec:theory}
	In this section, we show that the local solution computed by alternating minimization algorithm is guaranteed to converge to the truth though the problem is highly nonconvex.
	Our target is to bound the distance between the estimated coefficients $\widehat{\bmC}$ and its true counterpart $\bmC$ when the network structure is correctly specified.
	The distance is referred to the tensor angles. For two tensors $\mathcal{U}, \mathcal{V}$ of the same shape, define the distance (angle) between them as $\dist^2(\mathcal{U},\mathcal{V})=1-\langle\mathcal{U},\mathcal{V}\rangle^2/\left(\|\mathcal{U}\|_F^2\|\mathcal{V}\|_F^2\right)$. Here we focus on rank-1 Deep Kronecker Network under linear model while our results can be extended to general cases.
	\begin{condition}\label{RIP}
		{\bf \it (Restricted Isometry Property)}: Let $\bmX_i$ be the observed image tensors. Suppose that for all $\bmB^r_{l}\in\mathbb{R}^{d_l\times p_l\times q_l}$, for all $l\in[L]$ and $r=1,2$, there exists a constant $\delta\in (0,1)$ such that
		\begin{equation}
			(1-\delta)\left\|\sum_{r=1}^{2}\bigotimes_{l=L}^1 \bmB_{l}^r\right\|^2_F \le \frac{1}{n}\sum_{i=1}^n\left\langle \bmX_i, \sum_{r=1}^2\bigotimes_{l=L}^1 \bmB_{l}^r\right\rangle^2 \le (1+\delta)
			\left\|\sum_{r=1}^{2}\bigotimes_{l=L}^1 \bmB_{l}^r\right\|^2_F.
		\end{equation}
	\end{condition}

Now we provide an overview of main theory, with details deferred to Supplementary Material.
	\begin{theorem}\label{th-MAIN}
		Suppose that model $y_i=\left\langle\bmX_i, \ \bmC\right\rangle +\eps_i$ holds with $\bmC=\bigotimes_{l=L}^1 \bmB_{l}$. 
		Assume Condition \hyperref[RIP]{1} with a small-enough constant $\delta$ and  $\|\bep\|_2\le c(1-\delta)\|\bmC\|_F/2$ for certain constant $c$. Suppose that the likelihood function (\hyperref[obj1]{4}) is solved using alternating minimization algorithm with a correctly specified network structure and a spectral initialization. 
		Let $\kappa <1$ be a contraction parameter, $\mu$ be the initialization error and $\tau=\sqrt{(1/n)\log n}$. 
		Then, after $t$ iterations, the distance between estimates  $\widehat{\bmC}^{(t)}$ and $\bmC$ is bounded with high probability that 
		\begin{equation} \label{th-main-1}
			\dist\left(\widehat{\bmC}^{(t)}, \bmC\right)\le c_1\kappa^{t}\mu+c_2\tau,
		\end{equation}
		where $c_1, c_2$ are constants.
		
	\end{theorem}

	\begin{remark} \label{th-MAIN-1}
The first term in the right of (\hyperref[th-main-1]{8}) could be viewed as the optimization error, and the second term is the statistical error. 
Theorem \hyperref[th-MAIN]{3} suggests that the optimization error decays geometrically, even if the objective function  (\hyperref[obj1]{4}) is highly nonconvex. After $t\ge t_0+\frac{\log(n^{-1}\log n)}{2\log (\kappa)}$ iterations, we have $\dist\left(\widehat{\bmC}^{(t)}, \bmC\right) \asymp \sqrt{(1/n)\log n}$ holds with high probability. 
\end{remark}

	\begin{remark} \label{th-MAIN-2}
	Due to the connection between Deep Kronecker Network and tensor regression, Theorem \hyperref[th-MAIN]{3} also works for tensor regression solved by block relaxation algorithm. The spectral initialization required by Theorem \hyperref[th-MAIN]{3} is essential, as it can be proved to be not far away from truth. See Supplementary Material for more details.
\end{remark}

\section{The ADNI analysis}\label{sec:real}
\begin{figure}[t]
	\centering
	\includegraphics[width=\textwidth]{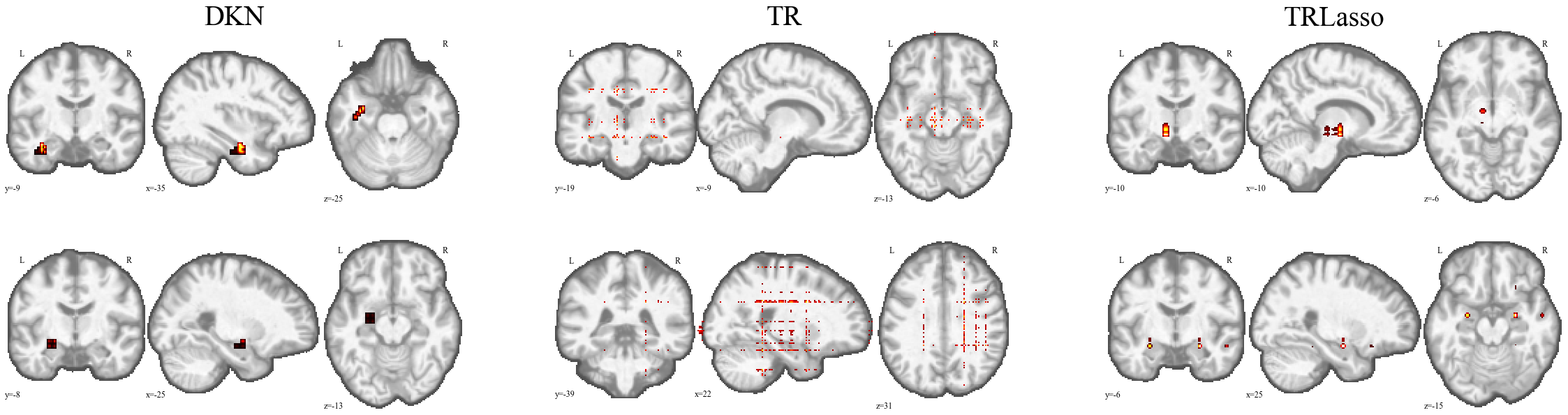}
	\caption{Detected regions in regression (top row) and classification (bottom row).}
	\label{fig:ADNI}
\end{figure}

In this section, we analyze Alzheimer's Disease (AD) with data collected from the Alzheimer’s Disease Neuroimaging Initiative (ADNI), a study designed to detect and track AD with clinical, genetic, imaging data, etc.
In ADNI analysis, we use T1-weighted MRI scans with two types of outcomes: i) binary outcomes for classification suggesting if participants have AD or not, and ii) continuous outcomes for regression suggesting the Mini-Mental State Examination (MMSE) score, a commonly used reference for the diagnosis of AD. 
After pre-processing, the images are represented as tensors of size $64^3$. We use the first two phases ADNI-1 and ADNI-GO as training and the third phase ADNI-3 as testing, resulting 417 subjects for training and 241 for testing.
Deep Kronecker Network is implemented under the deepest possible (6-layer) model with factors of size $2^3$ and ranks tuned by BIC, compared with three competing methods including CNN, tensor regression and tensor regression with Lasso penalty.
We report the prediction results of four methods in Table \hyperref[tab:ADNI]{1} and plot estimated coefficients in Fig. \hyperref[fig:ADNI]{3}.

By Table \hyperref[tab:ADNI]{1} and Fig. \hyperref[fig:ADNI]{4}, Deep Kronecker Network not only achieves the best prediction performance, but also detects the most precise region. Also, we note that the regions detected by Deep Kronecker Network in classification and regression are consistent, both around the hippocampus. In medical literature, hippocampus has been proved to be associated with AD, e.g. \cite{dubois2016preclinical}. Therefore, our findings are in line with existing medical literature.

\begin{table}[t]
	\def~{\hphantom{0}}
	\tbl{Results of the ADNI analysis. The best-performing method is marked with an asterisk.}{
		\begin{tabular}{ccccccc}
			Task&Criterion &   DKN  & TR     & TRLasso & CNN \\
			Regression            & RMSE      & *0.2258  & 0.2627 & 0.2557  & 0.2909 \\
			Classification        & Accuracy  & *79.25\% & 66.80\% & 76.76\%  & 78.01\%    
	\end{tabular}}
	\label{tab:ADNI}
\end{table}

\section*{Acknowledgment}
This work was funded in part by the Hong Kong RGC Grant ECS 21313922 and GRF 17301123.

\bibliographystyle{biometrika}
\bibliography{reference}

\section*{Supplementary material}\label{sec:supp}
\noindent In the supplementary material, we provide details for computation, theoretical results, numerical studies,  proofs and more discussions. The supplementary material is organized as follows: we introduce essential notations in Section \ref{sec:notation}. In Section \ref{sec:computation}, we introduce the alternating minimization algorithm for DKN computation. We in Section \ref{sec:structure} discuss the network structure and tuning parameter selection. Section \ref{sec:sims} contains comprehensive simulation studies on regression and classification. Section \ref{sec:real} provides more details on the MRI analysis from ADNI. In Section \ref{sec:thms}, we provide additional theoretical results. Finally, Section \ref{sec:proof} contains proofs for the theorems.

\setcounter{equation}{0}
\def\theequation{S\arabic{equation}}
\setcounter{figure}{0}
\def\thefigure{S\arabic{figure}}
\setcounter{table}{0}
\def\thetable{S\arabic{table}}
\setcounter{section}{0}
\def\thesection{S\arabic{section}}
\setcounter{proposition}{0}
\def\theproposition{S\arabic{proposition}}
\setcounter{algo}{0}
\def\thealgo{S\arabic{algo}}
\setcounter{remark}{0}
\def\theremark{S\arabic{remark}}
\setcounter{theorem}{0}
\def\thetheorem{S\arabic{theorem}}
\setcounter{lemma}{0}
\def\thelemma{S\arabic{lemma}}
\setcounter{corollary}{0}
\def\thecorollary{S\arabic{corollary}}
\setcounter{condition}{0}
\def\thecondition{S\arabic{condition}}

\section{Notation}\label{sec:notation}
{\bf Notation:} For $k\in\mathbb{N}$, let $[k] = \{1, \ldots, k\}$. We use calligraphic letters $\bmA$, $\bmB$ to denote tensors, including matrices, bold lower-case letters $\ba$, $\bb$ to denote vectors. We let $\tvec(\cdot)$ be the vectorization operator and $\tvec^{-1}_{(\cdot)}(\cdot)$ be its inverse with the subscripts subjecting the matrix/tensor size. For example, $\tvec^{-1}_{(d,p,q)}(\cdot)$ stands for transforming a vector of dimension $dpq$ to a tensor of dimension $d\times p\times q$. We let $\langle\cdot,\cdot\rangle$ to denote inner product, $\otimes$ to denote Kronecker product. 

We use square brackets around the indices to denote the entries of tensors. For example, suppose that $\bmA\in\mathbb{R}^{n_1\times n_2\times n_3\times n_4}$ is a four-order tensor. Then the entries of $\bmA$ is denoted as $\bmA_{[i_1],[i_2],[i_3],[i_4]}$. For simplicity, we may omit the square brackets when all indices are considered separate, i.e., $\bmA_{i_1,i_2,i_3,i_4}=\bmA_{[i_1],[i_2],[i_3],[i_4]}$. By forming indices together, we obtain lower order tensors. For example, a three-order tensor can be obtained by forming the first two indices together, with entries denoted by $\bmA_{[i_1i_2],[i_3],[i_4]}$. Here the grouped index $[i_1i_2]$ is equivalent to the linear index $i_1+n_1(i_2-1)$. Grouping the last three indices together results to a matrix (two-order tensor) with entries $\bmA_{[i_1],[i_2i_3i_4]}$, where the index $[i_2i_3i_4]$ denotes $i_2+n_2(i_3-1)+n_2n_3(i_4-1)$. When all the indices are grouped together, we obtain the vectorization of $\bmA$, also denoted as $\tvec(\bmA)$, with entries $\bmA_{[i_1i_2i_3i_4]}$.

\section{Computation}\label{sec:computation}
In this section, we propose an alternating minimization algorithm to solve DKN. The algorithm is illustrated for tensor images $\bmX_i\in\mathbb{R}^{d\times p\times q}$. 
We shall first consider the computation of DKN with a fixed structure, i.e., given factor number $L$, rank $R$ and factor sizes $(d_l, p_l, q_l)$, $l=1,\ldots, L$. The determination of network structure will be discussed in Section \ref{sec:structure}.

We need a few more notations to get started. Let $\bb^r_{l}=\tvec(\bmB^r_{l})\in\mathbb{R}^{d_lp_lq_l}$ be the vectorization of $\bmB^r_{l}$ for $l=1,\ldots, L$, $r=1,\ldots, R$. Let
\bes
\barB_{l}=\left[\bb^1_{l}, \bb^2_{l},\ldots, \bb^R_{l}\right]\in\mathbb{R}^{(d_lp_lq_l)\times R}, \ \ \barb_l=\tvec\left(\barB_l\right)
\ees
be the combined matrix of $\bb_l^{r}$ over different ranks and its vectorized version, respectively. Moreover, let $\bmB_{(:l)}^r$ and $\bmB_{(l:)}^r$  be the product of factors as below,
\bes
\bmB_{(:l)}^r=\bigotimes_{k=L}^l \bmB_k^r,  \ \ \bmB_{(:L+1)}^r=1, \ \ \bmB_{(l:)}^r=\bigotimes_{k=l}^1 \bmB_k^r,  \  \ \bmB_{(0:)}^r=1,  \ \  l=1,\ldots, L.
\ees
Further let $ \bb_{(:l)}^r$ and $\bb_{(l:)}^r$ be the vectorized version of $ \bmB_{(:l)}^r$ and $\bmB_{(l:)}^r$, respectively,
\bes
\bb_{(:l)}^r=\tvec\left(\bmB_{(:l)}^r\right), \ \ \ \bb_{(l:)}^r=\tvec\left(\bmB_{(l:)}^r\right), \ \ \ l=1,\ldots, L.
\ees
Finally, define the combined matrices of $\bb_{(:l)}^r$ and $\bb_{(l:)}^r$ over different ranks
\bes
\barB_{(:l)}&=&\left[\bb^1_{(:l)}, \bb^2_{(:l)},\ldots, \bb^R_{(:l)}\right], \ \ \   \barb_{(:l)}=\tvec\left(\barB_{(:l)}\right),
\cr\barB_{(l:)}&=&\left[\bb^1_{(l:)}, \bb^2_{(l:)},\ldots, \bb^R_{(l:)}\right], \ \ \ \barb_{(l:)}=\tvec\left(\barB_{(l:)}\right). 
\ees

Now we introduce a tensor reshaping operator. Let $\bmC\in\mathbb{R}^{d\times p\times q}$, and $d'$, $p'$, $q' \in\mathbb{R}_+$  that could be divided by $d$, $p$ and $q$ respectively. Let $(d'',p'',q'')=(d/d', p/p',q/q')$.
Define the operator $\mR_{(d', p', q')}: \mathbb{R}^{d\times p\times q}\rightarrow \mathbb{R}^{(d'p'q')\times (d''p''q'')}$ be a mapping from $\bmC$ to 
\begin{align*}\label{Rmat}
	\mR_{(d', p', q')}(\bmC)=&\Big[\tvec(\bmC_{1,1,1}^{d'',p'',q''}),\ldots, \tvec(\bmC_{1,1,q'}^{d'',p'',q''}),\ldots, \tvec(\bmC_{1,p',1}^{d'',p'',q''}),\ldots, \tvec(\bmC_{1, p',q'}^{d'',p'',q''}),\ldots, \cr
	&\ \ \tvec(\bmC_{d',1,1}^{d'',p'',q''}),\ldots, \tvec(\bmC_{d',1,q'}^{d'',p'',q''}),\ldots, \tvec(\bmC_{d',p',1}^{d'',p'',q''}),\ldots,\tvec(\bmC_{d', p',q'}^{d'',p'',q''})\Big]^\T.
\end{align*}
where $\bmC_{j,k,l}^{d'',p'',q''}$ is the $(j,k,l)$-th block of $\bmC$ of size $d''\times p''\times q''$.
A key property of the operator $\mathcal{R}$ is that for any tensor Kronecker product $\bmA\otimes \bmB$,
\bel{R_property}
\mR_{(d', p', q')}(\bmA\otimes \bmB) = \tvec(\bmA)\left[\tvec(\bmB)\right]^\T.
\eel
Given above definitions, we have the following Proposition.
\begin{proposition}\label{computation}
	Let $\tbX_i\left(\bb^r_{(:l+1)}, \ \bb^r_{(l-1:)}\right)$ be a function of $\bb^r_{(:l+1)}$ and $\bb^r_{(l-1:)}$,
	\begin{eqnarray*}
		&&\tbX_i\left(\bb^r_{(:l+1)}, \ \bb^r_{(l-1:)}\right) \cr
		&=&\mR_{(d_l,p_l,q_l)}\left(\tvec^{-1}_{\left(d_{(l:)},p_{(l:)},q_{(l:)}\right)}\left(\left[\bb^r_{(:l+1)}\right]^\T\mR_{\left( d_{(:l+1)}, p_{(:l+1)}, q_{(:l+1)}\right)}(\bX_i)\right)\right)\bb^r_{(l-1:)}.
	\end{eqnarray*}
	where we denote $d_{(l:)} = \prod_{j=l}^1 d_j$, $d_{(:l+1)} = \prod_{j=l+1}^L d_j$. The same notations are also used for $p$ and $q$.
	Furthermore, let $\barX_{i} (\barb_{(:l+1)}, \ \barb_{(l-1:)})$ be a function of $\barb_{(:l+1)}$ and $\barb_{(l-1:)}$,
	\bes
	\barX_{i} \left(\barb_{(:l+1)}, \ \barb_{(l-1:)}\right)=\left[\tbX_i\left(\bb^1_{(:l+1)}, \ \bb^1_{(l-1:)}\right),\ldots, \tbX_i\left(\bb^R_{(:l+1)}, \ \bb^R_{(l-1:)}\right)\right].
	\ees
	Then we have
	\begin{equation}\label{lm1-3}
		\left\langle\bmX_i, \sum_{r=1}^R\bigotimes_{l=L}^1 \bmB_l^r\right\rangle= \left[\tvec\left(\barX_{i} \left(\barb_{(:l+1)}, \ \barb_{(l-1:)}\right)\right)\right]^\T \barb_{l}.
	\end{equation}
	As a consequence, the loss function $\ell\left(\bmB^1_{1},\ldots, \bmB^R_{L}\right)$ could be written as 	
	\begin{align}\label{lm1-4}
		& \ell\left(\barb_{l}, \ \barb_{(:l+1)}, \ \barb_{(l-1:)}\right)
		\cr=&\sum_{i=1}^n\psi\Big(\Big[\tvec\Big(\barX_{i}\left(\barb_{(:l+1)},  \barb_{(l-1:)}\right)\Big)\Big]^\T\barb_{l}\Big)-y_i\Big[\tvec\Big(\barX_{i}\left(\barb_{(:l+1)}, \barb_{(l-1:)}\right)\Big)\Big]^\T \barb_{l}.
	\end{align}
	That is to say, given $\barb_{(:l+1)}$ and $\barb_{(l-1:)}$, the new $\widehat{\barb}_{l}$ can be updated by standard GLM estimation. With updated $\widehat{\barb}_{l}$ , we further have new
	\bel{lm1-5}
	&&\widehat{\barB}_{l}=\tvec_{(d_lp_lq_l, R)}^{-1}\left(\widehat{\barb}_{l}\right),
	\cr &&\hbb^r_{l}=\left[\widehat{\barB}_{l}\right]_{r,\cdot}, \  \text{the r-th column of }  \widehat{\barB}_{l},
	\cr && \widehat{\bmB}^r_{l} = \tvec_{(d_l, p_l,q_l)}^{-1}(\hbb^r_{l}),
	\cr && \hbb^r_{(l:)}=\tvec\left(\widehat{\bmB}^r_{l}\otimes \left[\widehat{\bmB}^r_{(l-1:)}\right]^{(old)}\right),
	\cr && \widehat{\barB}_{(l:)}=\left[\hbb^1_{(:l)}, \hbb^2_{(:l)},\ldots, \hbb^R_{(:l)}\right],
	\cr && \widehat{\barb}_{(l:)}=\tvec\left(\widehat{\barB}_{(l:)}\right).
	\eel
	\end{proposition}

	Proposition \ref{computation} suggests that the DKN could be solved by an alternating minimization algorithm with $\barb_{1},\ldots, \barb_{L}$ updated iteratively. 
	To implement the alternating minimization algorithm, the initializations of $\hbb_{(:2)},\hbb_{(:3)},\ldots,\hbb_{(:L)}$ are needed. They could be obtained by singular value decompositions as below:
	\bel{ini2}
	&&\widehat{\barb}_{(:l)}^{(0)}=\tvec\left(\widehat{\barB}_{(:l)}^{(0)}\right), \ \ \ \widehat{\barB}_{(:l)}^{(0)}=\left(\left[\hbb^1_{(l:)}\right]^{(0)}, \left[\hbb^2_{(l:)}\right]^{(0)},\ldots, \left[\hbb^R_{(l:)}\right]^{(0)}\right), \cr
	&&\left[\hbb_{(:l)}^r\right]^{(0)}=\text{SVD}_{u, r}\left(\sum_{i=1}^n y_i\mR_{\left(d_{(:l)},p_{(:l)},q_{(:l)}\right)}(\bmX_i)\right), \ \  l=2,3,\ldots, L,
	\eel
	where $\text{SVD}_{u,r}(\bX)$ denote the $r$-th top left singular vector of $\bX$.  We summarize the alternating minimization algorithm in Algorithm \ref{alg} below.
	
	\begin{algo}\label{alg}
		Alternating Minimization Algorithm for DKN
		\begin{tabbing}
			\qquad \enspace Input:  $\by_i$ and $\bmX_i$, $i=1,\ldots, n$.\\
			\qquad \enspace Initialization: $\left[\widehat{\barb}_{(:l)}^r\right]^{(0)}$ is obtained by (\ref{ini2}).\\ 
			\qquad \enspace For $t$ in $0, 1, \ldots, T-1$ \\
			\qquad \qquad  For $l$ in $1, 2, \ldots, L$\\
			\qquad \qquad \quad \quad $\widehat{\barb}_{l}^{(t+1)} \leftarrow\argmin_{\barb_l}  \ell\left(\barb_{l}, \ \widehat{\barb}_{(:l+1)}^{(t)}, \ \widehat{\barb}_{(l-1:)}^{(t+1)}\right)$, where $\ell(\cdot,\cdot,\cdot)$ defined in (\ref{lm1-4}). \\
			\qquad \qquad \quad \quad $\widehat{\barb}_{(l:)}^{(t+1)}$ updated by (\ref{lm1-5}). \\
			\qquad \qquad  For $l$ in $L, (L-1),\ldots, 1$\\
			\qquad \qquad \quad \quad $\widehat{\barB}_{l}^{(t+1)}\leftarrow\tvec_{(d_lp_lq_l, R)}^{-1}\left(\widehat{\barb}_{l}^{(t+1)}\right)$; \\
			\qquad \qquad \quad \quad $\left[\hbb^r_{l}\right]^{(t+1)}\leftarrow\left[\widehat{\barB}_{l}^{(t+1)}\right]_{r,\cdot}$ \\
			\qquad \qquad \quad \quad $\left[\widehat{\bmB}^r_{l}\right]^{(t+1)} \leftarrow \tvec_{(d_l, p_l,q_l)}^{-1}\left(\left[\hbb^r_{l}\right]^{(t+1)}\right)$ \\
			\qquad \qquad \quad \quad $\left[\widehat{\bmB}^r_{(:l)}\right]^{(t+1)}\leftarrow \left[\widehat{\bmB}^r_{(:l+1)}\right]^{(t+1)}\otimes \left[\widehat{\bmB}^r_{l}\right]^{(t+1)}$; \\
			\qquad \qquad \quad \quad $\left[\hbb^r_{(:l)}\right]^{(t+1)} \leftarrow \tvec\left(\left[\widehat{\bmB}^r_{(:l)}\right]^{(t+1)}\right)$ \\
			\qquad \qquad $\widehat{\barb}_{(:l)}^{(t+1)}\leftarrow \tvec\left(\left[\hbb^1_{(:l)}\right]^{(t+1)}, \left[\hbb^2_{(:l)}\right]^{(t+1)}, \ldots, \left[\hbb^R_{(:l)}\right]^{(t+1)}\right)$; \\
			
			\qquad \enspace Output $\widehat{\barb}_{1}^{(T)},\ldots,\widehat{\barb}_{L}^{(T)}$.
		\end{tabbing}
	\end{algo}

		\begin{remark} 
	Algorithm \ref{alg} could be viewed as in integration of a two-step procedure: 1) reshape the original images to obtain $\mathcal{T}(\bmX_i)$, and 2) implement tensor regression, such as block relaxation algorithm in \cite{zhou2013tensor}, on the reshaped images. 
	We shall emphasis that the reshaping step is crucial and it leads to different performances of DKN and TR.
\end{remark}

\section{Network Structure: Depth vs Width}\label{sec:structure}
In a convolutional neural network, or general deep neural network, the structure usually need to be carefully tuned in order to achieve the optimal prediction power. In particular, how the depth and width of a neural network would affect its prediction power has been intensively studied in the literature, to list a few, \citet{raghu2017expressive, lu2017expressive, tan2019efficientnet}.
Similarly, it is also of a concern in DKN how to find an optimal structure. In this subsection, we provide a general guidance on the determination of DKN structure.

To implement a DKN, the depth $L$, width $R$ and the filter sizes $\bmB_l^r$, i.e., $(d_l,p_l, q_l)$, need to be determined. Although they could all be treated as tuning parameters, 
we argue that it is not necessary to tune them all.

First, we note that for any given  $L$ and $(d_l,p_l, q_l)$, there exists a corresponding $R$ such that any tensor of size $\left(\prod_{l=1}^Ld_l,\ \prod_{l=1}^Lp_l, \ \prod_{l=1}^Lq_l\right)$ could be approximated. Such a result could be seen by relating KPD with CP decomposition. In other words, it is not necessary to tune the depth $L$ and filter sizes $(d_l,p_l, q_l)$ carefully.

Second, a deeper DKN is usually preferred. Recall that DKN is designed for image analysis under limited sample sizes. A deepest DKN allows us to achieve maximized dimension reduction. For example, suppose the images of concern are of size $(d,p,q)=(256,256,256)$. If we consider a 8-layer DKN with all the filters  are $2\times2\times 2$, then the total number of unknown parameters in a rank-R DKN is $R*64 (=2^3*8)$. As a comparison, the unknown parameter number in 2-layer, rank-$R$, filters size $(16,16,16)$ DKN is $R*8192 (=16^3*2)$. Certainly, a larger $R$ is possibly needed in a deeper DKN in order to achieve a better expressive power. But still, the benefit of depth is tremendous. In our simulation and real data analysis later, we stick to the deepest possible DKN.

Third, given $L$ and $(d_l,p_l, q_l)$, it is possible to design an information criterion to choose the rank $R$. For example, we may minimize the Bayesian Information Criterion (BIC)
\bel{bicc}
\text{BIC}(R)= 2\ell\left(\widehat{\bmB}_1^{1},\ldots, \widehat{\bmB}_L^{R}\right) + \left(R\sum_{l=1}^L d_lp_lq_l\right) \log n
\eel
In practice, we find that a relatively low rank model (e.g., $R=1,2,3$) in many cases would already produce desired estimation accuracy and prediction power. Therefore, we usually suggest to implement DKN from low-rank models. 


\section{Simulation studies}\label{sec:sims}
In this section, we conduct comprehensive simulation studies to demonstrate the prediction and coefficients estimation performance of DKN. We consider both regression and classification tasks, which are subject to a linear model and a logistic model respectively. Formally,

1. $y_i=\langle\bmX_i, \bmC\rangle+\ep_i, \quad \ep_i\sim\mathcal{N}(0,1)$

2. $y_i \sim \text{Ber}(\pi_i), \quad \text{logit}(\pi_i) = \langle\bmX_i, \bmC\rangle$

The simulation is conducted under different signal shapes, signal intensities and sample sizes. Specifically, we fix the image sizes at $128\times 128$,
but consider two different sample sizes $n=500,1000$. Each entry of image  $\bmX_i$ is generated from i.i.d. Gaussian $\mathcal{N}(0,1)$ distribution.

We consider four different coefficients matrices $\bmC$, including two sparse and two quasi-sparse coefficient matrices. Under both sparse and quasi-sparse cases, we consider two types of signal shapes: one circle and two circles. For the one-circle signal, the true signal is a circle centered at $(40, 88)$ with radius 10. While for the two-circle signal, the circles are centered at $(24, 40), \ (72, 88)$ respectively and both with radius 8. Under sparse case, $\bmC_{i,j}=1$ if $(i,j)$ falls in the signal region, and $\bmC_{i,j}=0$ otherwise. Under quasi-sparse case, $\bmC_{i,j}\sim \mathcal{N}(1,1)$ when $(i,j)$ falls in the signal region, and $\bmC_{i,j}\sim \mathcal{N}(0.1,0.1)$ otherwise. Apparently, quasi-sparse case could mimic real data applications better as it allows small perturbation beyond the signal region. We plot four coefficients matrices in the first column of Figure \ref{fig:1000} below. 


The DKN is implemented under the deepest possible model. That is to say, with images of size $128\times 128\ (=2^7\times 2^7)$, the number of layers is maximized to be $L=7$ and the sizes of factors $\bB_l^r$ are minimized to be $d_l=p_l=2$. We vary the rank of DKN and use the BIC in (\ref{bicc}) to select the optimal one. Note that none of four coefficient matrices could be exactly written in the form of $\sum_{r=1}^R\bigotimes_{l=L}^1 \bB_{l}^r$ with $R=1, 2, 3$. In other words, we are considering a mis-specified setting that is not in favor of DKN models. 

We compare the performance of DKN with four competing methods, namely, low-rank matrix regression \citep[LRMR,][]{zhou2014regularized}, tensor regression \citep[TR,][]{zhou2013tensor} and tensor regression with Lasso regularization (TRLasso), and CNN. The LRMR imposes a nuclear norm on the coefficients so that the produced coefficients matrix is of low-rank. The TR and TRLasso are designed for tensor input, but could still be adapted for matrix images. As for CNN, we consider a typical structure with two convolutional layers (followed by max-poolings) and two fully connected layers. In convolutional layers, the kernel size is $5\times5$ and stride size is $1\times1$. 
The activation function is ReLU and batch normalization is applied. 
We evaluate the coefficients estimation and prediction performance of different methods. The estimation performance is measured by the root mean squared error (RMSE): $\|\widehat{\bmC}-\bmC\|_F/\sqrt{dp}$. To evaluate the prediction performance, we independently generate an additional $n_{test}=n/4$ samples. Then the prediction error is measured by RMSE for regression task, i.e. $\sqrt{(1/n_{test})\sum_{i=1}^{n_{test}}(\hat{y}_i^{test}-y_i^{test})^2}$ and accuracy for classification task, i.e. $(1/n_{test})\sum_{i=1}^{n_{test}}I\{\hat{y}_i^{test}=y_i^{test}\}$. Note that for CNN, only the prediction performance could be evaluated, as there is no recoverable estimated coefficients. The simulation results are averaged over 100 independent repetitions and reported in Table \ref{tab:reg} and \ref{tab:cla}. In each task, the best results are marked as green. In addition, we plot the estimated coefficients of different methods in Figure \ref{fig:1000}.

By Table \ref{tab:reg}, \ref{tab:cla} and Figure \ref{fig:1000}, it is clear that DKN performs extremely competitive across a large range of settings in both regression and classification tasks, and the former is even better. In particular, when the sample size is small ($n=500$), the DKN approach demonstrates dominating performance with the smallest estimation and prediction errors.
As discussed earlier, the DKN is designed for such a low-sample size scenario, which commonly exists in medical imaging analysis. The simulation study further validated the advantage of DKN under such a setting.

When the sample size increases to $n=1000$, we could find other methods showing advantages under some cases, such as the TRLasso under the sparse one-circle case. However, the DKN approach is still the best performer in general, especially under the quasi-sparse case. Compared to the sparse case, quasi-sparse coefficients matrices are more difficult to be recovered. But the DKN could still locate the most influential regions and achieve the best estimation accuracy. On the other hand, we shall note that when sample size increases, the improvement of DKN in two-circle case is more than that in one-circle case. That is because BIC tends to selected DKN models with larger ranks in the two-circle case. It further suggests the benefits of including more DKN terms when the sample size is large.
\\

\begin{figure}[t]
	\centering
	\includegraphics[width=0.9\textwidth]{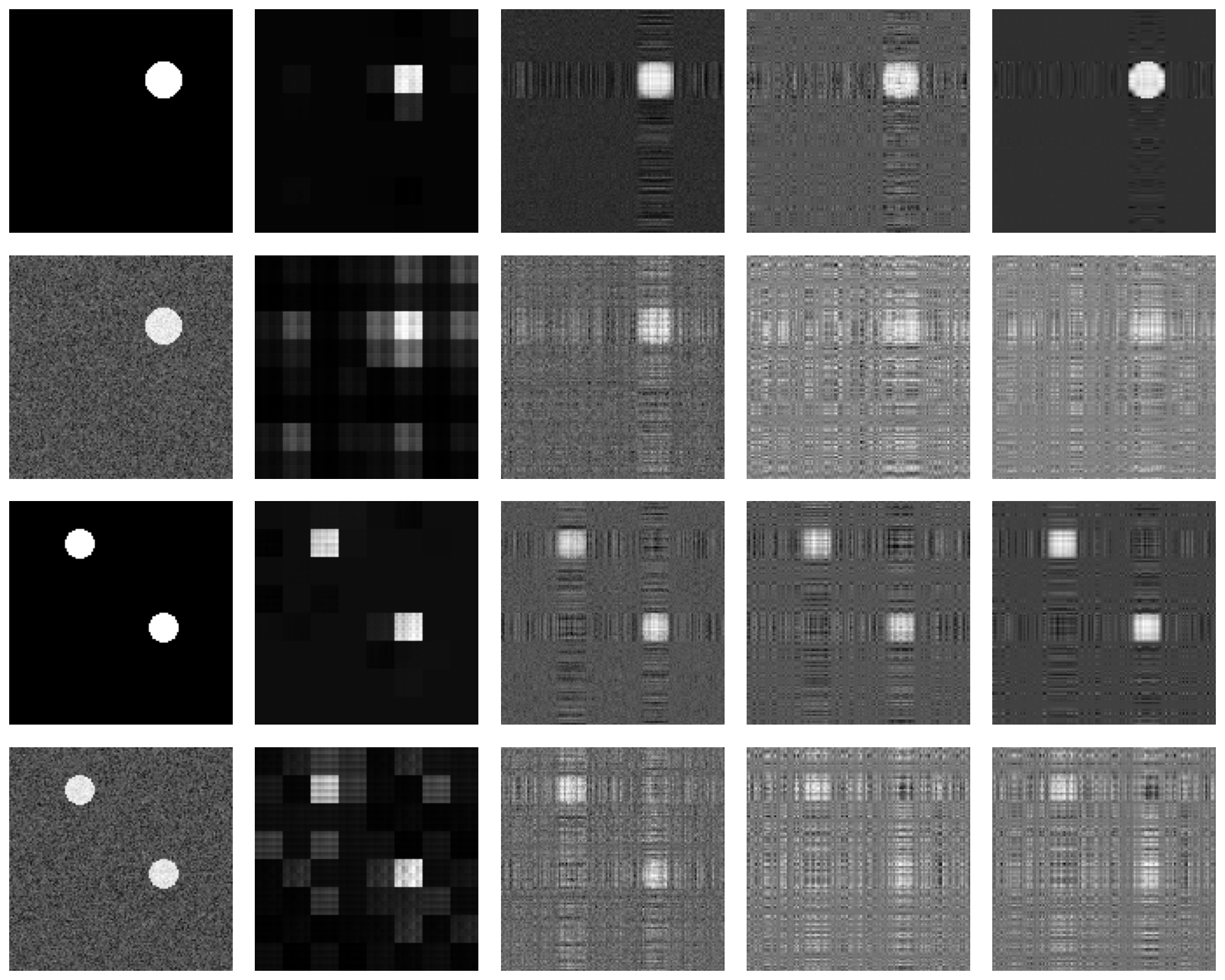}
	\caption{An illustration of the estimated coefficients matrices under $n=1000$ in regression case. 
		Rows from top to bottom: sparse one-circle, quasi-sparse one-circle, sparse two-circle, quasi-sparse two-circle. Columns from left to right: true signal, DKN, LRMR, TR, TRLasso.}
	\label{fig:1000}
\end{figure}

\begin{table}[t]
	\def~{\hphantom{0}}
	\tbl{Simulation study on regression.}{%
	\begin{tabular}{cccccc}
		\multirow{2}{*}{One-circle} & Sparsity    & \multicolumn{2}{c}{Sparse}    & \multicolumn{2}{c}{Quasi-Sparse} \\
		& Sample size & 500           & 1000          & 500             & 1000           \\
		\multirow{4}{*}{Estimation} & DKN         & 0.037 (0.002) & 0.035 (0.001) & 0.140 (0.003)   & 0.137 (0.005)  \\
		& LRMR        & 0.113 (0.004) & 0.072 (0.003) & 0.189 (0.003)   & 0.161 (0.003)  \\
		& TR          & 0.251 (0.052) & 0.105 (0.010) & 0.437 (0.046)   & 0.307 (0.017)  \\
		& TRLasso     & 0.197 (0.066) & 0.042 (0.024) & 0.295 (0.062)   & 0.258 (0.023)  \\
		&             &               &               &                 &                \\
		\multirow{5}{*}{Prediction} & DKN         & 10.06 (0.725) & 9.83 (0.529)  & 18.35 (1.266)   & 18.20 (1.138)  \\
		& LRMR        & 14.60 (0.943) & 9.32 (0.558)  & 24.27 (1.523)   & 20.72 (0.963)  \\
		& TR          & 31.85 (6.862) & 13.48 (1.494) & 56.03 (6.650)   & 39.10 (2.735)  \\
		& TRLasso     & 25.32 (8.949) & 5.57 (3.076)  & 38.19 (8.232)   & 32.90 (3.444)  \\
		& CNN         & 14.42 (0.989) & 11.25 (0.640) & 22.58 (1.362)   & 20.15 (1.066)  \\
		\multicolumn{6}{l}{} \\
		\multirow{2}{*}{Two-circle} & Sparsity    & \multicolumn{2}{c}{Sparse}    & \multicolumn{2}{c}{Quasi-Sparse} \\
		& Sample size & 500           & 1000          & 500             & 1000           \\
		\multirow{4}{*}{Estimation} & DKN         & 0.064 (0.011) & 0.056 (0.012) & 0.159 (0.007)   & 0.150 (0.012)  \\
		& LRMR        & 0.154 (0.002) & 0.118 (0.004) & 0.209 (0.002)   & 0.184 (0.004)  \\
		& TR          & 0.349 (0.034) & 0.152 (0.025) & 0.481 (0.038)   & 0.354 (0.026)  \\
		& TRLasso     & 0.270 (0.061) & 0.120 (0.023) & 0.302 (0.041)   & 0.300 (0.030)  \\
		\multicolumn{6}{l}{} \\
		\multirow{5}{*}{Prediction} & DKN         & 14.48 (3.269) & 13.24 (3.583) & 22.04 (1.823)   & 20.73 (2.276)  \\
		& LRMR        & 19.87 (1.145) & 15.28 (0.806) & 26.87 (1.643)   & 23.74 (1.107)  \\
		& TR          & 44.52 (4.880) & 19.30 (3.219) & 61.26 (6.706)   & 45.18 (3.976)  \\
		& TRLasso     & 34.44 (7.949) & 15.42 (3.048) & 38.93 (6.365)   & 38.41 (4.057)  \\
		& CNN         & 17.13 (1.221) & 13.92 (0.660) & 23.98 (1.670)   & 21.87 (1.369) 
	\end{tabular}}
	\centering
	\label{tab:reg}
\end{table}

\begin{table}[!htbp]
	\def~{\hphantom{0}}
	\tbl{Simulation study on classification.}{%
	\begin{tabular}{cccccc}
		\multirow{2}{*}{One-circle} & Sparsity    & \multicolumn{2}{c}{Sparse}    & \multicolumn{2}{c}{Quasi-Sparse} \\
		& Sample size & 500           & 1000          & 500             & 1000           \\
		\multirow{4}{*}{Estimation} & DKN         & 0.129 (0.007) & 0.128 (0.001) & 0.195 (0.005)   & 0.192 (0.001)  \\
		& LRMR        & 0.146 (0.000) & 0.146 (0.000) & 0.204 (0.000)   & 0.204 (0.000)  \\
		& TR          & 0.178 (0.030) & 0.116 (0.013) & 0.242 (0.017)   & 0.183 (0.007)  \\
		& TRLasso     & 0.132 (0.002) & 0.134 (0.006) & 0.203 (0.002)   & 0.196 (0.002)  \\
		\multicolumn{6}{l}{} \\
		\multirow{5}{*}{Accuracy} & DKN         & 0.758 (0.104) & 0.813 (0.024) & 0.643 (0.123)   & 0.738 (0.040)  \\
		& LRMR        & 0.516 (0.048) & 0.575 (0.038) & 0.505 (0.044)   & 0.549 (0.035)  \\
		& TR          & 0.564 (0.069) & 0.779 (0.027) & 0.529 (0.058)   & 0.651 (0.035)  \\
		& TRLasso     & 0.762 (0.061) & 0.843 (0.026) & 0.534 (0.057)   & 0.691 (0.038)  \\
		& CNN         & 0.634 (0.049) & 0.737 (0.030) & 0.593 (0.049)   & 0.674 (0.032)  \\
		\multicolumn{6}{l}{} \\
		\multirow{2}{*}{Two-circle} & Sparsity    & \multicolumn{2}{c}{Sparse} & \multicolumn{2}{c}{Quasi-Sparse} \\
		& Sample size & 500           & 1000          & 500             & 1000           \\
		\multirow{4}{*}{Estimation} & DKN         & 0.154 (0.012) & 0.150 (0.010) & 0.211 (0.004)   & 0.208 (0.003)  \\
		& LRMR        & 0.166 (0.000) & 0.166 (0.000) & 0.218 (0.000)   & 0.218 (0.000)  \\
		& TR          & 0.214 (0.011) & 0.157 (0.003) & 0.259 (0.011)   & 0.209 (0.004)  \\
		& TRLasso     & 0.164 (0.002) & 0.166 (0.000) & 0.218 (0.001)   & 0.214 (0.001)  \\
		\multicolumn{6}{l}{} \\
		\multirow{5}{*}{Accuracy} & DKN         & 0.689 (0.108) & 0.768 (0.068) & 0.628 (0.096)   & 0.706 (0.040)  \\
		& LRMR        & 0.505 (0.047) & 0.535 (0.033) & 0.504 (0.044)   & 0.527 (0.035)  \\
		& TR          & 0.526 (0.049) & 0.603 (0.035) & 0.520 (0.048)   & 0.595 (0.041)  \\
		& TRLasso     & 0.546 (0.061) & 0.790 (0.052) & 0.520 (0.051)   & 0.611 (0.042)  \\
		& CNN         & 0.626 (0.047) & 0.728 (0.029) & 0.612 (0.047)   & 0.675 (0.033)  
	\end{tabular}}
	\centering
	\label{tab:cla}
\end{table}

\subsection{Additional Simulations for nonlinear DKN}
In this subsection, we implemented nonlinear DKN and compared its performance with linear DKN. We consider the same simulation setting as before. Specifically, we consider sample sizes $n=500, 1000$ and four different coefficients matrices $\bC$, namely sparse one-circle, sparse two-circle, quasi-sparse one-circle, and quasi-sparse two-circle.

Table \ref{tab:nonlinear} shows the prediction error of DKN and nonlinear DKN under different signal coefficients and sample sizes. 
The nonlinear DKN is considered with the same network structure as its linear version and implemented using a popular deep learning optimization algorithm, Adam \citep{kingma2014adam}.
The nonlinear activation function is chosen as Leaky-ReLU. 
By Table  \ref{tab:nonlinear}, it is clear that the linear DKN is advantageous across all the settings. On one hand, the setting is more favorable to DKN as the outcome is generated through linear model. 
On the other hand, SGD algorithms like Adam are designed for large sample problems and their performance could be affected when the sample size is limited. 
It is an interesting future direction to study the practical value of nonlinear activation in DKN, including its applied scenario, network structure, optimization, etc.

\begin{table}[t]
	\def~{\hphantom{0}}
	\tbl{Prediction error of DKN and Nonlinear DKN under different simulation settings.}{%
	\begin{tabular}{ccccc}
		Sparsity      & \multicolumn{2}{c}{Sparse}    & \multicolumn{2}{c}{Quasi-Sparse} \\
		Sample size   & 500           & 1000          & 500             & 1000           \\
		Shape         & \multicolumn{4}{c}{One-circle}                                   \\
		DKN           & 10.06 (0.725) & 9.83 (0.529)  & 18.35 (1.266)   & 18.20 (1.138)  \\
		Nonlinear DKN & 16.16 (1.572) & 15.46 (1.253) & 28.13 (6.239)   & 26.85 (3.947)  \\
		Shape         & \multicolumn{4}{c}{Two-circle}                                   \\
		DKN           & 14.48 (3.269) & 13.24 (3.583) & 22.04 (1.823)   & 20.73 (2.276)  \\
		Nonlinear DKN & 18.44 (1.404) & 17.31 (2.190) & 26.49 (2.052)   & 27.07 (1.731) 
	\end{tabular}}
	\centering
	\label{tab:nonlinear}
\end{table}

\subsection{Adam vs AMA for DKN computation}

In this subsection, we implement DKN using Adam and compare its performance with alternating minimization algorithm (AMA). A comprehensive simulation study suggests that Algorithm \ref{alg} is advantageous under the low-sample-size scenario.

Specifically, we fix the image size to be $128\times 128$ and vary the number of observations $n=500,1000,2000$. The true coefficients matrix represents a sparse circle signal as in the manuscript. We fit the deepest rank-1 DKN (number of layers $L=7$ and factor sizes $d_l=p_l=2$) using Algorithm \ref{alg}  and Adam separately. Table \ref{adam} below reports the estimation error (root mean squared error), prediction error and computation time of two algorithms.

By Table \ref{adam},  it is clear that the estimation and prediction performance of Adam is close to Algorithm \ref{alg}  when sample size is large (n=2000), but the performance would deteriorate when sample size decreasing. As a comparison, our alternating algorithm adapts to low sample size well and produces much more stable estimation. The stochastic gradient descent (SGD) algorithms like Adam are designed for large sample problems and their performance could be affected when the sample size is small.  
In the literature, \cite{liu2019variance} studied the convergence issue of Adam. They showed that a root cause of the convergence issue is the undesired large variance of adaptive learning rate, which in fact caused by limited amount of training samples. 

\begin{table}[t]
	\def~{\hphantom{0}}
	\tbl{A comparison between Alternating Minimization Algorithm (Algorithm \ref{alg} ) and Adam.}{%
		\begin{tabular}{ccccc}
			\multirow{2}{*}{One-Circle} & Sparsity     & \multicolumn{3}{c}{Sparse}                    \\
			& Sample size  & 500           & 1000          & 2000          \\
			\multirow{2}{*}{Estimation} & Algorithm \ref{alg} & 0.036 (0.002) & 0.035 (0.001) & 0.034 (0.001) \\
			& Adam         & 0.054 (0.023) & 0.052 (0.021) & 0.034 (0.001) \\
			\multirow{2}{*}{Prediction} & Algorithm \ref{alg} & 9.952 (0.818) & 9.791 (0.462) & 9.890 (0.392) \\
			& Adam         & 14.03 (5.188) & 13.29 (4.221) & 9.891 (0.395) \\
			\multirow{2}{*}{Time}       & Algorithm \ref{alg} & 3.517 (0.234) & 5.770 (0.059) & 13.64 (1.629) \\
			& Adam         & 10.76 (7.193) & 17.42 (20.03) & 41.77 (26.91)
	\end{tabular}}
	\label{adam}
\end{table}

Regarding the computation time, we are surprised to find that the Algorithm \ref{alg}  demonstrated a dominating performance compared to Adam. The major reason is that Algorithm \ref{alg}  could converge within a few (around 10) iterations, while Adam may need hundreds of epochs despite with comparable GPUs. In summary, when samples are limited as in medical imaging analysis, the alternating minimization algorithm provides a better option compared to DL frameworks not only in estimation accuracy, but also in computational efficiency. 

\section{The ADNI analysis}\label{sec:real}
In this section, we use MRI data to analyze the Alzheimer's Disease (AD), with data collected from the Alzheimer’s Disease Neuroimaging Initiative (ADNI). The ADNI is a study designed to detect and track Alzheimer's disease with clinical, genetic, imaging data, etc. We refer to the website \textit{https://adni.loni.usc.edu/} for more details.

In the ADNI analysis, we use MRI data to analyze two types of outcomes: i) binary outcomes suggesting whether the participants have AD or not, and ii) continuous outcomes suggesting the Mini-Mental State Examination (MMSE) score of participants. The MMSE score is designed to assess the cognitive impairment of a patient. By  \cite{tombaugh1992mini}, an MMSE score falling in the region of [24, 30], [19, 23], [10, 18] and [0, 9] suggests no, mild, moderate and severe cognitive impairment, respectively.
Therefore, the MMSE score could also be viewed as a reference for the diagnosis of Alzheimer's disease. In other words, these two outcomes considered here are highly correlated.

\begin{figure}[t]
	\centering
	\includegraphics[width=0.7\textwidth]{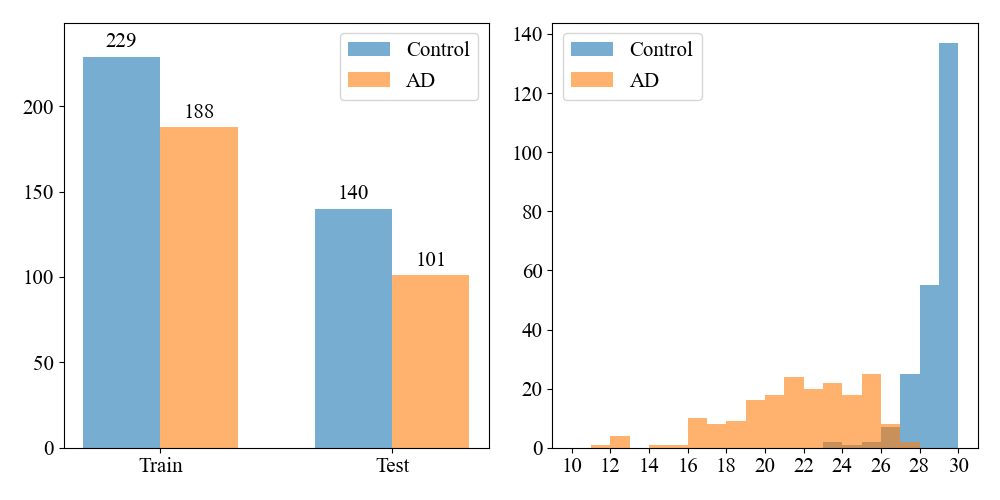}
	\caption{Left: summary of AD vs control in the training and test set; Right: distribution of the MMSE scores, with AD and controls are marked by different colors.}
	\label{fig:dist}
\end{figure}

The ADNI has four phases of study until today: ADNI-1, ADNI-GO, ADNI-2 and ADNI-3. As ADNI-3 is still ongoing, our analysis focuses on the first three phases. 
Specifically, we use data in ADNI-1 and ADNI-GO phase as training set while data in ADNI-2 phase as test set. The training set and test set contains 417 and 241 subjects, respectively. The distributions of AD status (for classification) and the outcome MMSE (for regression) are plotted in Figure \ref{fig:dist}.

Each participant in the analysis is involved with a T1-weighted MRI scan. The T1-weighted MRI scan were carefully preprocessed before analysis.  A standard pipeline proceeds as follows: spatial adaptive non-local means (SANLM) denoising \citep{manjon2010adaptive}, resampling, bias-correction, affine-registration and unified segmentation, skull-stripping and cerebellum removing \citet{ashburner2005unified}. It follows that local intensity correction and spatial normalization (into the Montreal Neurological Institute (MNI) atlas space). Each T1-weighted MRI scan is thereby processed into a tensor of size $113\times 137\times 113$. To improve analytical efficiency, we first resize each image into a smaller tensor and conduct zero-padding. The finally obtained images are represented as tensors of size $64\times64\times64$.

\subsection{Regression analysis for MMSE}
In this subsection, we use the MRI data to predict the MMSE score. As discussed before, the MMSE score is a continuous outcome ranging from 0 to 30.
Normal people usually has an MMSE score close to 30 (mean 28.82, s.d. 1.02 in our dataset). While for AD patients, the mean and standard deviation are 21.63 and 3.25, respectively.

As in the simulation, we implement the deepest DKN model. Specifically, we consider a 6-layer DKN with factors $\bmB_l^r$ of size $2\times 2 \times 2$. We still consider DKN models with the rank selected by BIC. Moreover, the performance of DKN is compared with TR, TRLasso and CNN. Note that we didn't implement the LRMR because it is unable to be generalized for tensor inputs. For CNN, we refer to the networks in \cite{hu2020brain}, which also studied MRI data using CNN. Specifically, we consider a network with two convolutional layers, two max-pooling layers and two fully-connected layers. In convolutional layers, the kernel size is $3\times3\times 3$ and the stride size is $1\times1\times 1$. In max-pooling layers, the kernel size is $2\times 2\times 2$. ReLU activation and batch normalization are applied additionally.

We report the test set prediction accuracy of different methods in Table \ref{tab:ADNI}. We also visualize the estimated coefficient tensors in Fig. \ref{fig:ADNI} except for CNN. We refer to the significantly non-zero positions in the coefficient tensor as the detected region. To illustrate the detected regions better, the coefficient tensors of DKN and TR are plotted after thresholding (more than 10 times of standard deviation).

By Table \ref{tab:ADNI}, DKN clearly performs best with the smallest prediction error. As a comparison, the CNN obtains the largest RMSE, suggesting a larger sample size is needed for it. Moreover, by Figure \ref{fig:ADNI}, the brain region detected by DKN (colored area) is indicating an area around the hippocampus, which has been shown to associated with AD in medical literature (to be discussed later).
While for TR and TRLasso, they failed to capture the region of hippocampus, resulting to compromised prediction accuracy. 

In the literature, the hippocampus has been proved to be associated with Alzheimer’s disease. For example, the early work of \cite{ball1985new} has attributed the decline of higher cognitive functions in AD to the hippocampus and proposed to name AD as a hippocampal dementia. \cite{dubois2016preclinical} revealed that AD would gradually destroy different areas of brain cells and hippocampus is one of the regions suffering the damage first. 
Therefore, we are able to claim that the findings of DKN is in line with existing medical literature. 


\subsection{Classification analysis for AD}


In this subsection, we conduct a binary classification analysis that uses MRI data to predict the participants' AD status. 
The training set contains 417 subjects with $45\%$ AD patients, while the test set contains 241 subjects with $42\%$ AD patients.

We employ the same DKN structure as in the regression analysis: the number of layers $L=6$, the factors $\bmB_l^r$ are of size $(2\times 2 \times 2)$ and Kronecker ranks selected by BIC. For DKN, TR and TRLasso, a logit link function is employed for such a binary classification task. While for CNN, we also use the same structure described in the regression analysis (two convolutional layer, two max-pooling and two fully connected layers), but with a soft-max output function for the classification problem.

The classification accuracy and region detection results are reported in Table \ref{tab:ADNI} and Fig. \ref{fig:ADNI} (right), respectively. We again observe that DKN achieved the highest classification accuracy. On the other hand, we note that although TRLasso performs a little worse, the TR without regularization performs the worst among all methods. 
In terms of region detection performance, we see that the brain areas detected by DKN and TRLasso are all located around hippocampus, but the TR again failed to capture such area. 



Combining the regression and classification analyses, we see that DKN is the only approach that is able to locate hippocampus under both cases. In conclusion, the DKN could not only achieve the best possible prediction accuracy under limited sample size, more importantly, it could also provide desired interpretability and help medical researchers understand imaging data better.

\section{Additional Theorems}\label{sec:thms}
In this section, we provide additional theoretical results. The section is divided into two parts. 
The first part contains the identifiability conditions of DKN. The second part provides details on the theoretical error bounds of DKN. 

\subsection{Identifiability conditions for $\bmB^r_{l}$}\label{sec:idf}
In general, when the structure of DKN, including the depth $L$, width $R$ and factor sizes $(d_l,p_l,q_l)$, $l=1,\ldots, L$, is unknown, the unknown tensors $\bmB^r_{l}$ are not identifiable. Therefore, we here focus on the case that the structure of DKN is given and derive the conditions under which the $\bmB^r_{l}$ are identifiable.

Before discussing the identifiability condition, we shall first realize two elementary indeterminacies of KPD, namely scaling and permutation. If a tensor $\bmC$ can be represented by KPD with target tensors $\bmB^r_{l}$, we use the notation $\bmC=[\![\bmB^r_{l}]\!]$ to refer this decomposition. Meanwhile, recall the notation $\bb^r_{l}=\tvec(\bmB^r_{l})$ and $\barB_{l}=\left[\bb^1_{l}, \bb^2_{l},\ldots, \bb^R_{l}\right]$. So we also use
$\bmC=[\![\bar{\bB}_{1},\ldots, \bar{\bB}_{L}]\!]$ to refer the same decomposition.
The scaling indeterminacy states that 
$\bmC=[\![\bmB^r_{l}]\!]=[\![\bmB^r_{l}\gamma_{l}^r]\!]$, when $\prod_{l=1}^L \gamma_{l}^r=1$, for all $r=1,\ldots, R$.
The permutation indeterminacy states that 
$\bmC=[\![\bar{\bB}_{1},\ldots, \bar{\bB}_{L}]\!]=[\![\bar{\bB}_{1} \Omega,\ldots, \bar{\bB}_{L}\Omega]\!]$,
where $\bOmega$ is certain permutation matrix. To avoid the two indeterminacies, we impose the following constraints. We first let $\lam_r=\prod_{l=1}^L\|\bmB^r_{l}\|_F$ to denote the $r$-th ``Kronecker eigenvalue (KE)'' in KPD. To address the scaling indeterminacy, we fix $\|\bmB_1^{r}\|_F=\lam_r$ and  $\|\bmB^r_{l}\|_F=1$ for $l=2,\ldots,L$ across all the terms $r=1,\ldots,R$. To address the permutation indeterminacy, we permute $\bmB^r_{l}$ such that $\lam_1\ge\cdots\ge\lam_R$ for all the layers $l=1,\ldots,L$.

Now we are ready to state the sufficient and necessary conditions for identification. 
\begin{theorem}\label{th-id}
	Suppose $\bmC$ has a KPD form $\bmC=\sum_{r=1}^R\bigotimes_{l=L}^1 \bmB_{l}^r$. Suppose that the configuration of DKN, including the depth $L$, width $R$ and block sizes $(d_l,p_l,q_l)$, $l=1,\ldots, L$, are correctly specified. Let $\bb^r_{l}=\tvec(\bmB^r_{l})$ and $\barB_{l}=\left[\bb^1_{l}, \bb^2_{l},\ldots, \bb^R_{l}\right]$. Then:
	\begin{itemize}
		\item[1.] (Sufficiency) The KPD is unique up to scaling and permutation if 
		\bes
		\sum_{l=1}^L K\left(\barB_{l}\right)\ge 2R+ L-1
		\ees
		where $K(\bM)$ is the $K$-rank of a matrix $\bM$, i.e., the maximum value K such that any K columns of $\bM$ are linearly independent. 
		\item[2.] (Necessity) If the KPD is unique up to scaling and permutation, then
		\bes
		\min_{l=1,\cdots L} \left\{\prod_{l'\neq l} \rank \left(\barB_{l}\right)\right\}\ge R
		\ees
	\end{itemize}
\end{theorem}
\begin{remark}
	When $L=2$, the KPD could be transformed into singular value decomposition (SVD). So we immediately have the sufficient and necessary condition for KPD:
	$\left\langle \bmB^{r_1}_{1}, \bmB^{r_2}_{1}\right\rangle=0$, $\left\langle \bmB^{r_1}_{2}, \bmB^{r_2}_{2}\right\rangle=0$, $ \forall r_1\neq r_2$.
	The KPD under $L=2$ has been intensively studied in the literature and it was usually named as Kronecker product singular value decomposition (KPSVD). We refer to \citet{van1993approximation} for more details.
\end{remark}
Theorem \ref{th-id} is built upon the existing theorem on CPD along with the connections between KPD and CPD. In particular, the sufficiency condition is based on \citet{sidiropoulos2000uniqueness}, while the necessity condition is based on \citet{liu2001cramer}.




\subsection{Theoretical error bounds}\label{sec:bounds}
In this subsection, we provide details on the theoretical error bounds of DKN. Specifically, we prove that the alternating minimization algorithm described in Section \ref{sec:computation} is able to guarantee the resulted coefficients $\hmC$ converge to the true $\bmC$ even though the problem is highly nonconvex.  For ease of presentation, we here focus DKN with rank 1 under the linear model setting, although our results could be extended to general R-term KPD. That is to say, we suppose the model is generated from
\bel{model00}
y_i=\left\langle\bmX_i, \ \bigotimes_{l=L}^1 \bmB_{l}\right\rangle +\eps_i.  
\eel
where $\eps_i$ are i.i.d. noises. Also, note that we omit the subscripts $r$ under the case $R=1$.  

	Our target is to bound the distance between the estimated coefficients $\widehat{\bmC}$ and its true counterpart $\bmC$ when the network structure is correctly specified.
Here the distance is referred to the tensor angles.	For any two tensors $\mathcal{U}, \mathcal{V}$ of the same shape, define the distance (angle) between $\mathcal{U}$ and $\mathcal{V}$ as $\dist^2(\mathcal{U},\mathcal{V})=1-\langle\mathcal{U},\mathcal{V}\rangle^2/\left(\|\mathcal{U}\|_F^2\|\mathcal{V}\|_F^2\right)$,
where $\|\cdot\|_F$ is the Frobenius norm. 


There are two assumptions needed to guarantee the convergence of DKN. One is the Restricted Isometry Property (RIP) introduced in the manuscript. The second is an initialization condition. 
But we shall first define a quantity related to the error term $\eps_i$. We first recall that $\bmB_{(:l)}$ and $\bmB_{(l:)}$ are respectively the product of factors from $L$ to $l$ and from $l$ to 1, and $\bb_{(:l)}=\tvec(\bmB_{(:l)})$ and $\bb_{(l:)}=\tvec(\bmB_{(l:)})$ are their vectorized version. We also recall the  transformation $\tbX_i\left(\bb_{(:l+1)}, \ \bb_{(l-1:)}\right)$ in Proposition \ref{computation}. Then, define
\bes
\tau_0=\sup\left\{\frac{1}{n}\Big\|\sum_{i=1}^n\eps_i\tbX_i\left(\bb_{(:l+1)}, \ \bb_{(l-1:)}\right)\Big\|_2, \ \|\bb_{(:l+1)}\|_2= \|\bb_{(l-1:)}\|_2=1,\  l=1,\ldots, L \right\}.
\ees
\begin{condition}\label{ini}
	{\bf \it (Initialization)} Let $\mu_l=\dist\left(\hbb_{(:l)}^{(0)}, \bb_{(:l)}\right)$  be the initial estimation error of the factor product, $l=2,3,\ldots, L$. Let $\mu=\max_{l=2}^L\{\mu_l\}$ be the maximum of $\mu_l$. Let $\delta$ be the contant in the RIP condition and $\tau_0$ as above.  Further let $\tau=(\tau_0/\|\bmC\|_F)(1-3\delta)^{-1}$, $\nu=\mu+3\delta/(1-3\delta)$ and $\eta=\mu/[\mu+\tau(\nu+1)/\nu]$. Suppose
	$\nu < (1+\eta)^{\frac{1}{L-1}}-1$.
\end{condition}
\begin{remark}
	Under a noiseless case $\eps_i=0$, $i=1,\ldots, n$, we have $\tau=0$ and thus $\eta=1$. As a consequence, the Condition 2 is reduced to $\nu<2^{1/(L-1)}-1$.
\end{remark}

The Condition \ref{ini} imposes a requirement for the initial error. The magnitude of the initial error shall be controlled by the noise level and RIP constant. In Theorem \ref{init} below, we show the Condition  \ref{ini} could be satisfied easily with an initialization in (\ref{ini2}).

Given the RIP and initialization condition, we are ready to state our main theory.
\begin{theorem}\label{th-main}
	(Non-Asymptotic) Suppose model (\ref{model00}) holds and Algorithm \ref{alg} is implemented under a correctly specified network structure. Assume that the images $\bmX_i$ satisfies RIP condition with constant $\delta$. Let $\mu_l=\dist\left(\hbb_{(:l)}^{(0)}, \bb_{(:l)}\right)$, $\mu=\max_{l=2}^L\{\mu_l\}$, $\nu=\mu + 3\delta/(1-3\delta)$, $\tau=(\tau_0/\|\bmC\|_F)(1-3\delta)^{-1}$ and $\kappa=(\nu+1)^L - (2\nu+1)$. Suppose the initialization Condition \ref{ini} holds. 
	Then, after t times iteration, the distance between $\widehat{\bmC}^{(t)}$ and $\bmC$ is bounded by
	\bel{th-main-1}
	\dist\left(\widehat{\bmC}^{(t)}, \bmC\right)\le c_1\kappa^{t}\mu
	+c_2\tau
	\eel
	where $c_1$ and $c_2$ are explicit constants: $c_1=(L-1)(1+\nu/\kappa)$ and $c_2=(1+\nu)^2/[\nu(1-\kappa)]+1$. 
\end{theorem}
The $\kappa$ in Theorem \ref{th-main} could be viewed as a contraction parameter and it is guaranteed to be less than 1 under Condition \ref{ini} for initialization. The first term in the RHS of (\ref{th-main-1}) could be viewed as the optimization error, while the second term is the statistical error.  
By Theorem \ref{th-main}, it is clear that the optimization error decays geometrically under the alternating minimization algorithm, even if the objective function is highly nonconvex. Moreover, when the error term $\eps$ is sub-Gaussian, the statistical error could be controlled by the probabilistic upper bound $\tau=\mathcal{O}_p\left(\sqrt{\frac{\log (n)}{n}}\right)$. 
As a consequence, we have the following corollary.
\begin{corollary}\label{cor-main}
	(Asymptotic) Suppose the conditions of Theorem \ref{th-main} hold. If the noise $\eps_i$ is sub-Gaussian, then when the sample size $n\rightarrow \infty$ and the times of iteration
	$t\ge t_0+\frac{\log(n^{-1}\log n)}{2\log (\kappa)}$,
	we have
	$\dist\left(\widehat{\bmC}^{(t)}, \bmC\right)\asymp \sqrt{\frac{\log (n)}{n}}$ holds with high probability, where $t_0$ is certain constant.
\end{corollary}

For CNN, it is difficult to guarantee that the computed solutions (by stochastic gradient descent or other algorithm) converge to the truth due to the non-convexity. But Theorem \ref{th-main} provides a different story for DKN. 
The key to prove Theorem \ref{th-main} is the following theorem. It guarantees that the approximation error in Theorem \ref{th-main} is decaying geometrically.

\begin{theorem}\label{lemmat}
	(Iteration) Suppose model (\ref{model00}) holds and Algorithm \ref{alg} is implemented under a correctly specified network structure. Assume that the images $\bmX_i$ satisfies RIP condition with constant $\delta$. Let $\mu_l=\dist\left(\hbb_{(:l)}^{(0)}, \bb_{(:l)}\right)$, $\mu=\max_{l=2}^L\{\mu_l\}$, $\nu=\mu + 3\delta/(1-3\delta)$ and $\tau=(\tau_0/\|\bmC\|_F)(1-3\delta)^{-1}$. Suppose the initialization Condition \ref{ini} holds. Then, for all $t=0, 1, \ \ldots$ and $l=1, \ldots, \ L$ we have
	\begin{align}
		\dist\left(\hbb_l^{(t+1)}, \bb_l\right) \le \nu\left[\dist\left(\hbb_{(l-1:)}^{(t+1)}, \bb_{(l-1:)}\right) + \dist\left(\hbb_{(:l+1)}^{(t)}, \bb_{(:l+1)}\right)\right]+\tau.  \label{central}
	\end{align}
	Note the special case for $l=0$ or $L+1$,
	$\dist\left(\hbb_{(0:)}^{(t)}, \bb_{(0:)}\right) =\dist\left(\hbb_{(:L+1)}^{(t)}, \bb_{(:L+1)}\right)=0$.
\end{theorem}
Theorem \ref{lemmat} could be proved by a carefully constructed power method. We refer to the Section \ref{sec:lemmas} for the proof of Theorem \ref{lemmat}. On the other hand, the Condition \ref{ini} for initialization is required in Theorem \ref{th-main} and Theorem \ref{lemmat}. Now we show that if the initialization is taken as in (\ref{ini2}), such a initialization condition could be satisfied easily.

\begin{theorem}\label{init}
	(Initialization)	
	Suppose model (\ref{model00}) holds and Algorithm \ref{alg} is implemented under a correctly specified network structure. Assume that the images $\bmX_i$ satisfies RIP with constant $\delta$. Assume the initialization is taken as (\ref{ini2}) and the noise term satisfies  $\|\bep\|_2\le c(1-\delta)\|\bmC\|_F/2$ for certain constant $c$. Then, 
	\bes
	\max_{l=2,\ldots L} \left\{\dist\left(\hbb_{(:l)}^{(0)}, \bb_{(:l)}\right)\right\} \le c(1+\delta) + \frac{\sqrt{\delta(1+\delta)}}{1-\delta}.
	\ees
\end{theorem}

\section{Proofs}\label{sec:proof}
We provide proofs  for Theorem \ref{th-main} to \ref{init}.  The proof of Theorem \ref{th-id} is omitted as explained before. This section is divided into two parts: the first subsection provides additional lemmas with proofs, the second subsection gives the proofs of main theorem. 
\subsection{Proof of lemmas}\label{sec:lemmas}
\begin{lemma}\label{lm-0}
	For any two vectors $\bu, \bv\in\mathbb{R}^d$, we have:
	\bes
	\dist^2(\bu,\bv)&\le& \left(\frac{\bu}{\|\bu\|_2}-\frac{\bv}{\|\bv\|_2}\right)^2
	\cr \dist^2(\bu,\bv)&\le&\frac{\|\bu-\bv\|_2^2}{\|\bu\|_2^2}
	\cr \dist^2(\bu,\bv)&\le&\frac{\|\bu-\bv\|_2^2}{\|\bv\|_2^2}
	\ees
	Moreover, for any vectors $\bu,\hbu, \bv, \hbv\in\mathbb{R}^d$,
	\bes
	\dist^2(\bu\bv^\T,\ \hbu\hbv^\T)&= &\dist^2(\bu,\hbu)^2 + \dist^2(\bv,\hbv) -\dist^2(\bu,\hbu) \dist^2(\bv,\hbv)
	\ees
	It further follows that
	\bes
	\dist^2(\bu\bv^\T,\ \hbu\hbv^\T)&\le &\dist^2(\bu,\hbu) + \dist^2(\bv,\hbv)
	\cr \dist(\bu\bv^\T,\ \hbu\hbv^\T)&\le& \dist(\bu,\hbu) + \dist(\bv,\hbv)
	\cr \dist(\bu\bv^\T,\ \hbu\hbv^\T)&\ge& \dist(\bu,\hbu)  \dist(\bv,\hbv)
	\ees
	Furthermore, for any matrices $\bU,\hbU\in\mathbb{R}^{d_1\times p_1}$ and $\bV,\hbV\in\mathbb{R}^{d_2\times p_2}$ 
	\bes
	\text{max}\left\{\dist(\bU,\hbU), \ \dist(\bV,\hbV)\right\}\le\dist(\bU\otimes\bV,\hbU\otimes\hbV)\le\dist(\bU, \hbU) + \dist(\bV, \hbV).
	\ees
	More generally, for any matrices $\bU_{k}, \ \hbU_{k}\in\mathbb{R}^{d_k\times p_k}$, $k=1,2\ldots,l$, denote $\bU_{(:)}=\bigotimes_{k=1}^l\bU_{k}$ and $ \hbU_{(:)}=\bigotimes_{k=1}^l\hbU_k$, we have
	\bes
	\max_{k=1,\ldots,l}\left\{\dist\left(\bU_{k},\hbU_{k}\right)\right\}\le\dist\left(\bU_{(:)}, \hbU_{(:)}\right)
	\le\sum_{k=1}^l\dist\left(\bU_{k}, \hbU_{k}\right).
	\ees
	
\end{lemma}
We omit the proof of Lemma \ref{lm-0} as it could be derived easily by algebra.
\hfill$\square$
\bigskip

\begin{lemma}\label{lm-rip}
	Suppose the RIP condition holds for $\bmX_i$. Then for any $\bmB_l^r$, $l=1,\ldots, L$, $r=1,2$, we have
	\bes
	\left|\frac{1}{n}\sum_{i=1}^n\left\langle \bmX_i, \bigotimes_{l=L}^1 \bmB_l^{1}\right\rangle \left\langle \bmX_i, \bigotimes_{l=L}^1 \bmB_l^{2}\right\rangle-\left\langle\bigotimes_{l=L}^1 \bmB_l^{1}, \bigotimes_{l=L}^1 \bmB_l^{2}\right\rangle\right|\le 3\delta\left\|\bigotimes_{l=L}^1 \bmB_l^{1}\right\|_F \left\|\bigotimes_{l=L}^1 \bmB_l^{2}\right\|_F
	\ees
\end{lemma}
\bigskip

\noindent{\bf Proof of Lemma \ref{lm-rip}.} Due to the RIP condition,
\bes
&& \frac{1}{n}\sum_{i=1}^n\left\langle \bmX_i, \ \  \sum_{r=1}^2\bigotimes_{l=L}^1 \bmB_l^{r}\right\rangle^2 \le (1+\delta)
\left\|\sum_{r=1}^{2}\bigotimes_{l=L}^1 \bmB_l^{r}\right\|^2_F
\cr \Rightarrow&&\frac{1}{n} \sum_{i=1}^n\left\langle \bmX_i, \bigotimes_{l=L}^1 \bmB_l^{1}\right\rangle^2 + \frac{1}{n}\sum_{i=1}^n\left\langle \bmX_i, \bigotimes_{l=L}^1 \bmB_l^{2}\right\rangle^2 +\frac{2}{n} \sum_{i=1}^n\left\langle\bmX_i, \bigotimes_{l=L}^1 \bmB_l^{1}\right\rangle \left\langle \bmX_i, \bigotimes_{l=L}^1 \bmB_l^{2}\right\rangle
\cr &&\le (1+\delta)\left(\left\|\bigotimes_{l=L}^1 \bmB_l^{1}\right\|^2_F + \left\|\bigotimes_{l=L}^1 \bmB_l^{2}\right\|^2_F\right)+2(1+\delta)\left\langle\bigotimes_{l=L}^1 \bmB_l^{1}, \bigotimes_{l=L}^1 \bmB_l^{2}\right\rangle
\cr \Rightarrow&&\frac{1}{n}\sum_{i=1}^n\left\langle \bmX_i, \bigotimes_{l=L}^1 \bmB_l^{1}\right\rangle \left\langle \bmX_i, \bigotimes_{l=L}^1 \bmB_l^{2}\right\rangle-\left\langle\bigotimes_{l=L}^1 \bmB_l^{1}, \bigotimes_{l=L}^1 \bmB_l^{2}\right\rangle
\cr &&\le \delta\left(\left\|\bigotimes_{l=L}^1 \bmB_l^{1}\right\|^2_F + \left\|\bigotimes_{l=L}^1 \bmB_l^{2}\right\|^2_F\right)+\delta\left\langle\bigotimes_{l=L}^1 \bmB_l^{1}, \bigotimes_{l=L}^1 \bmB_l^{2}\right\rangle
\cr\Rightarrow&&\frac{1}{n}\sum_{i=1}^n\left\langle \bmX_i, \bigotimes_{l=L}^1 \bmB_l^{1}\right\rangle \left\langle \bmX_i, \bigotimes_{l=L}^1 \bmB_l^{2}\right\rangle -\left\langle\bigotimes_{l=L}^1 \bmB_l^{1}, \bigotimes_{l=L}^1 \bmB_l^{2}\right\rangle\cr &&\le \delta\left(\left\|\bigotimes_{l=L}^1 \bmB_l^{1}\right\|^2_F + \left\|\bigotimes_{l=L}^1 \bmB_l^{2}\right\|^2_F\right)+\delta\left\|\bigotimes_{l=L}^1 \bmB_l^{1}\right\|_F \left\|\bigotimes_{l=L}^1 \bmB_l^{2}\right\|_F
\ees
Furthermore, we note that the last inequality still holds if we replace $\bigotimes_{l=L}^1 \bmB_l^{1}$ by $\lam\bigotimes_{l=L}^1 \bmB_l^{1}$ and replace $\bigotimes_{l=L}^1 \bmB_l^{2}$ by $(1/\lam)\bigotimes_{l=L}^1 \bmB_l^{2}$. Optimizing the RHS with $\lam$, we get
\bes
\frac{1}{n}\sum_{i=1}^n\left\langle \bmX_i, \bigotimes_{l=L}^1 \bmB_l^{1}\right\rangle \left\langle \bmX_i, \bigotimes_{l=L}^1 \bmB_l^{2}\right\rangle -\left\langle\bigotimes_{l=L}^1 \bmB_l^{1}, \bigotimes_{l=L}^1 \bmB_l^{2}\right\rangle\le 3\delta\left\|\bigotimes_{l=L}^1 \bmB_l^{1}\right\|_F \left\|\bigotimes_{l=L}^1 \bmB_l^{2}\right\|_F.
\ees
The other side of the inequality could be proved similarly. This completes the proof.\hfill$\square$
\bigskip
\begin{lemma}\label{lm-delta}
	Suppose that $\|\bmB_l\|=1$ and $\|\widehat{\bmB}_l\|=1$, $l=l,\ldots, L$. Define $\hbSigma_{(l)}^{(t)}$ and $\bSigma_{(l)}^{(t)}$ respectively as
	\bes
	&&\hbSigma_{(l)}^{(t)} = (1/n)\sum_{i=1}^n \left[\hbZ_{i}^{(l)}\right]^{(t)}\hbb_{(l-1:)}^{(t+1)}\left[\hbb_{(l-1:)}^{(t+1)}\right]^\T\left(\left[\hbZ_{i}^{(l)}\right]^{(t)}\right)^\T \cr
	&&\bSigma_{(l)}^{(t)} = (1/n)\sum_{i=1}^n \left[\hbZ_{i}^{(l)}\right]^{(t)}\hbb_{(l-1:)}^{(t+1)}\left[\hbb_{(l-1:)}^{(t+1)}\right]^\T\left(\bZ_{i}^{(l)}\right)^\T .
	\ees
	Then we have
	\bes \left\|\left(\hbSigma_{(l)}^{(t)}\right)^{-1} \left(\langle\hbb^{(t)}_{:l+1},\bb_{:l+1} \rangle\hbSigma_{(l)}^{(t)}-\bSigma_{(l)}^{(t)}\right)\right\|_2\le\frac{3\delta}{1-3\delta}\dist\left(\hbb_{(:l+1)}^{(t)},\bb_{(:l+1)}\right).
	\ees
\end{lemma}
\bigskip

\noindent{\bf Proof of Lemma \ref{lm-delta}.} The minimum eigenvalue of $\hbSigma_{(l)}^{(t)}$ is given by
\begin{align}\label{lm-delta-1}
	&\lam_{\min}\left(\hbSigma_{(l)}^{(t)}\right) \cr =& \min_{\|\bb_l\|_2=1}\bb_l^\T\hbSigma_{(l)}^{(t)} \bb_l \cr
	=& (1/n)\sum_{i=1}^n\bb_l^\T \left[\hbZ_{i}^{(l)}\right]^{(t)}\hbb_{(l-1:)}^{(t+1)}\left[\hbb_{(l-1:)}^{(t+1)}\right]^\T\left(\left[\hbZ_{i}^{(l)}\right]^{(t)}\right)^\T \bb_l\cr =&(1/n)\min_{\|\bb_l\|_2=1}\sum_{i=1}^n\tr\left(\left[\hbZ_{i}^{(l)}\right]^{(t)}\hbb_{(l-1:)}^{(t+1)}\bb_l^\T\right) \tr\left(\left[\hbZ_{i}^{(l)}\right]^{(t)}\hbb_{(l-1:)}^{(t+1)}\bb_l^\T\right) \cr
	=&(1/n)\sum_{i=1}^n\left\langle \bmX_i, \widehat{\bmB}_{(:l+1)}\otimes\bmB_l\otimes\widehat{\bmB}_{(l-1:)}\right\rangle^2
	\cr \ge& 1-3\delta.
\end{align}
where the last inequality holds by Lemma \ref{lm-rip}. Further consider the term
\begin{eqnarray}\label{lm-delta-2}
	&&\left\|\langle \hbb_{(:l+1)}^{(t)},\bb_{(:l+1)}\rangle \hbSigma_{(l)}^{(t)}-\bSigma_{(l)}^{(t)}\right\|_2 \cr
	&=&\max_{\|\bu\|_2=1,\|\bv\|_2=1}\bu^\T\left(\langle \hbb_{:l+1}^{(t)},\bb_{:l+1}\rangle \hbSigma_{(l)}^{(t)}-\bSigma_{(l)}^{(t)}\right)\bv \cr
	&=&\max_{\|\bu\|_2=1,\|\bv\|_2=1}\sum_{i=1}^n\Big\{\langle\hbb_{(:l+1)}^{(t)},\bb_{(:l+1)}\rangle\bu^\T\left[\hbZ^{(l)}_i\right]^{(t)}\hbb_{(l-1:)}\hbb_{(l-1:)}^\T\left(\left[\hbZ^{(l)}_i\right]^{(t)}\right)^\T\bv \cr
	&& \ \ \ \ \ \ \ \ \ \ \ \ \ \ \ \ \ \ \ \ \ \  -\bu^\T\left[\hbZ^{(l)}_i\right]^{(t)}\hbb_{(l-1:)}\hbb_{(l-1:)}^\T\left(\bZ^{(l)}_i\right)^\T\bv \Big\}\cr
	&=&\max_{\|\bu\|_2=1,\|\bv\|_2=1}\sum_{i=1}^n\left\langle \bmX_i, \ \widehat{\bmB}_{(:l+1)}\otimes\bU\otimes \widehat{\bmB}_{(l-1:)}\right\rangle
	\cr &&\ \ \ \ \ \ \ \ \ \ \ \ \ \ \ \ \ \ \times\left\langle \bmX_i, \ \left(\langle\hbb_{(:l+1)}^{(t)},\bb_{(:l+1)}\rangle\widehat{\bmB}_{(:l+1)}\right)\otimes\bV\otimes \widehat{\bmB}_{(l-1:)}\right\rangle \cr
	&& \ \ \ \ \ \ \ \ \ \ \ \ \ \ \ \ \ \ -\left\langle \bmX_i, \ \widehat{\bmB}_{(:l+1)}\otimes\bU\otimes \widehat{\bmB}_{(l-1:)}\right\rangle \left\langle \bmX_i, \ \bmB_{(:l+1)}\otimes\bV\otimes \widehat{\bmB}_{(l-1:)}\right\rangle \cr
	&=&\max_{\|\bu\|_2=1,\|\bv\|_2=1}\sum_{i=1}^n\left\langle \bmX_i, \ \widehat{\bmB}_{(:l+1)}\otimes\bU\otimes \widehat{\bmB}_{(l-1:)}\right\rangle 
	\cr &&\ \ \ \ \ \ \ \ \ \ \ \ \ \ \ \ \ \ \times\left\langle \bmX_i, \ \left(\langle\hbb_{(:l+1)}^{(t)},\hbb_{(:l+1)}\rangle\widehat{\bmB}_{(:l+1)}-\bmB_{(:l+1)}\right)\otimes\bV\otimes \widehat{\bmB}_{(l-1:)}\right\rangle \cr
	&\le&3\delta\sqrt{1-\left\langle\hbb_{(:l+1)}^{(t)},\bb_{(:l+1)}\right\rangle^2} \cr
	&=& 3\delta \ \dist\left(\hbb_{(:l+1)}^{(t)},\bb_{(:l+1)}\right)
\end{eqnarray}
The last inequality holds by Lemma \ref{lm-rip}. Combining (\ref{lm-delta-1}) and (\ref{lm-delta-2}), we have
\bes 
\left\|\left(\hbSigma_{(l)}^{(t)}\right)^{-1} \left(\langle\hbb^{(t)}_{:l+1},\bb_{:l+1} \rangle\hbSigma_{(l)}^{(t)}-\bSigma_{(l)}^{(t)}\right)\right\|_2\le\frac{3\delta}{1-3\delta}\dist\left(\hbb_{(:l+1)}^{(t)},\bb_{(:l+1)}\right).
\ees
\hfill$\square$
\bigskip
\begin{lemma}\label{t+1}
	Suppose model (\ref{model00}) holds and Algorithm \ref{alg} is implemented under a correctly specified network structure. Let $\mu$ be the maximum of initial error, $\nu=\mu + 3\delta/(1-3\delta)$ and $\kappa=(\nu+1)^L - (2\nu+1)$.
	If 
	\bes
	\dist\left(\hbb_{(l-1:)}^{(t+1)},\bb_{(l-1:)}\right) \le \mu, \ \ \  \dist\left(\hbb_{(:l+1)}^{(t)},\bb_{(:l+1)}\right) \le \mu,
	\ees
	then
	\begin{equation}\label{central_l}
		\dist\left(\hbb_l^{(t+1)}, \bb_l\right)\le\nu\left(\dist\left(\hbb_{(l-1:)}^{(t+1)}, \bb_{(l-1:)}\right) + \dist\left(\hbb_{(:l+1)}^{(t)}, \bb_{(:l+1)}\right)\right)+\tau.
	\end{equation}
\end{lemma}
\bigskip

\noindent{\bf Proof of Lemma \ref{t+1}.} Lemma \ref{t+1}  provides the central inequality in our proof. For ease of presentation, we prove Lemma \ref{t+1} for matrix images. The tensor case follows the same way. First recall that $\tbX_i\left(\bb_{(:l+1)},  \bb_{(l-1:)}\right)$ is defined as
\bes
\tbX_i\left(\bb_{(:l+1)},  \bb_{(l-1:)}\right)=\mR_{(d_l,p_l)}\left(\tvec^{-1}_{\left(d_{(l:)}, p_{(l:)}\right)}\left(\bb_{(:l+1)}^\T\mR_{\left(d_{(:l+1)}, p_{(:l+1)}\right)}(\bX_i)\right)\right)\bb_{(l-1:)}.
\ees
Then we denote 
\bes
&&\bZ_i^{(l)} = \mR_{(d_l,p_l)}\left(\tvec^{-1}_{\left(d_{(l:)}, p_{(l:)}\right)}\left(\bb_{(:l+1)}^\T\mR_{\left(d_{(:l+1)}, p_{(:l+1)}\right)}(\bX_i)\right)\right), \cr
&&\left[\hbZ_i^{(l)}\right]^{(t)} = \mR_{(d_l,p_l)}\left(\tvec^{-1}_{\left(d_{(l:)}, p_{(l:)}\right)}\left(\left[\hbb_{(:l+1)}^{(t)}\right]^{\T}\mR_{\left(d_{(:l+1)}, p_{(:l+1)}\right)}(\bX_i)\right)\right) .
\ees
Moreover, let
\bes
&&\hbSigma_{(l)}^{(t)} = (1/n)\sum_{i=1}^n \left[\hbZ_{i}^{(l)}\right]^{(t)}\hbb_{(l-1:)}^{(t+1)}\left[\hbb_{(l-1:)}^{(t+1)}\right]^\T\left(\left[\hbZ_{i}^{(l)}\right]^{(t)}\right)^\T \cr
&&\bSigma_{(l)}^{(t)} = (1/n)\sum_{i=1}^n \left[\hbZ_{i}^{(l)}\right]^{(t)}\hbb_{(l-1:)}^{(t+1)}\left[\hbb_{(l-1:)}^{(t+1)}\right]^\T\left(\bZ_{i}^{(l)}\right)^\T \cr
&&\bTheta_{(l)}^{(t)} = (1/n)\sum_{i=1}^n \left[\hbZ_{i}^{(l)}\right]^{(t)}\hbb_{(l-1:)}^{(t+1)}\bb_{(l-1:)}^\T\left(\bZ_{i}^{(l)}\right)^\T
\ees
Without loss of generality, suppose $\hbb_{(l-1:)}^{(t+1)}$ and $\hbb_{(:l+1)}^{(t)}$ are normalized such that  $\hbb_{(l-1:)}^{(t+1)}=\hbb_{(:l+1)}^{(t)}=1$. 

Denote $\lambda=\|\bmC\|_F$. Given $\hbb_{(l-1:)}^{(t+1)}$ and $\hbb_{(:l+1)}^{(t)}$, we need to estimate $\hbb_l^{(t+1)}$. Denote the (normalized) estimates as $\hbb_l^{(t+1)}$ and its estimated norm as $\hlambda^{(t+1)}$. Then,
\bes
&&\hlambda^{(t+1)}\hbb_l^{(t+1)}\cr&=&\left(\frac{1}{n}\hbX_{(l)}^\T\hbX_{(l)}\right)^{-1}\frac{1}{n}\hbX_{(l)}^\T\left( \bX_{(l)}\lambda\bb_l + \bep\right) \cr
&=&\left(\hbSigma_{(l)}^{(t)}\right)^{-1}\left(\bTheta_{(l)}^{(t)}\lambda\bb_l +\frac{1}{n}\hbX_{(l)}^\T\bep\right)\cr
&=&\langle \hbb^{(t+1)}_{(l-1:)},\bb_{(l-1:)} \rangle \lambda\bb_l-\left(\hbSigma_{(l)}^{(t)}\right)^{-1}\left(\langle \hbb^{(t+1)}_{(l-1:)},\bb_{(l-1:)} \rangle\hbSigma_{(l)}^{(t)}-\bTheta_{(l)}^{(t)}\right)\lambda\bb_l +\frac{1}{n} \left(\hbSigma_{(l)}^{(t)}\right)^{-1}\hbX_{(l)}^\T\bep\cr
&=&\langle \hbb^{(t+1)}_{(l-1:)},\bb_{(l-1:)} \rangle \langle\hbb^{(t)}_{(:l+1)},\bb_{(:l+1)} \rangle\lambda\bb_l \cr
&&-\left(\hbSigma_{(l)}^{(t)}\right)^{-1}\left(\langle\hbb^{(t+1)}_{(l-1:)},\bb_{(l-1:)} \rangle \left[\langle\hbb^{(t)}_{(:l+1)},\bb_{(:l+1)}\rangle\hbSigma_{(l)}^{(t)}-\bSigma_{(l)}^{(t)}\right]\right)\lambda\bb_l \cr 
&&-\left(\hbSigma_{(l)}^{(t)}\right)^{-1}\left(\langle \hbb^{(t+1)}_{(l-1:)},\bb_{(l-1:)} \rangle\bSigma_{(l)}^{(t)}-\bTheta_{(l)}^{(t)}\right)\lambda\bb_l +\frac{1}{n} \left(\hbSigma_{(l)}^{(t)}\right)^{-1}\hbX_{(l)}^\T\bep, 
\ees
where $\hbX_{(l)}$ and $\bX_{(l)}$ are respectively
\bes
&&\hbX_{(l)}=\left[\tbX_1^\T\left(\hbb_{(:l+1)}^{(t)},  \hbb_{(l-1:)}^{(t+1)}\right) ,\cdots, \tbX_n^\T\left(\hbb_{(:l+1)}^{(t)},  \hbb_{(l-1:)}^{(t+1)}\right) \right]^\T,
\cr&&\bX_{(l)}=\left[\tbX_1^\T\left(\bb_{(:l+1)},  \bb_{(l-1:)}\right) ,\cdots, \tbX_n^\T\left(\bb_{(:l+1)},  \bb_{(l-1:)}\right) \right]^\T.
\ees
It then follows that
\bel{pf-1-1}
&&\frac{\|\hlambda^{(t+1)}\hbb^{(t+1)}_l-\lambda\bb_l\|_2}{\lambda} \cr
&\le&\underbrace{\left|1-\langle \hbb^{(t+1)}_{(l-1:)},\bb_{(l-1:)} \rangle\langle\hbb^{(t)}_{(:l+1)},\bb_{(:l+1)} \rangle\right|^{1/2}}_{A1} \cr
&&+\underbrace{\left\|\left(\hbSigma_{(l)}^{(t)}\right)^{-1}\left(\langle\hbb^{(t+1)}_{(l-1:)},\bb_{(l-1:)} \rangle\left[\langle\hbb^{(t)}_{(:l+1)},\bb_{(:l+1)} \rangle\hbSigma_{(l)}^{(t)}-\bSigma_{(l)}^{(t)}\right]\right)\right\|_2}_{A2}
\cr &&+\underbrace{\left\|\left(\hbSigma_{(l)}^{(t)}\right)^{-1}\left(\langle \hbb^{(t+1)}_{(l-1:)},\bb_{(l-1:)} \rangle\bSigma_{(l)}^{(t)}-\bTheta_{(l)}^{(t)}\right)\right\|_2}_{A3} 
\cr &&+ \underbrace{\frac{\left\|\left(\hbSigma_{(l)}^{(t)}\right)^{-1}\hbX_{(l)}^\T\bE\right\|_2}{\lambda}}_{\text{A4}}. 
\eel
We will bound A1 to A4 separately. For A1, we have
\bes
&&\left|1-\langle \hbb^{(t+1)}_{(l-1:)},\bb_{(l-1:)} \rangle \langle\hbb^{(t)}_{(:l+1)},\bb_{(:l+1)} \rangle\right|
\cr&\le&\left|1-\langle \hbb^{(t+1)}_{(l-1:)},\bb_{(l-1:)} \rangle^2 \langle\hbb^{(t)}_{(:l+1)},\bb_{(:l+1)} \rangle^2\right| \cr
&\le&\left|1-\langle \hbb^{(t+1)}_{(l-1:)},\bb_{(l-1:)} \rangle^2 \right| + \left|1- \langle\hbb^{(t)}_{(:l+1)},\bb_{(:l+1)} \rangle^2\right| \cr
&=&\dist^2(\hbb_{(l-1:)}^{(t+1)},\bb_{(l-1:)}) + \dist^2(\hbb_{(:l+1)}^{(t)},\bb_{(:l+1)}) \cr 
&\le&\mu\dist(\hbb_{(l-1:)}^{(t+1)},\bb_{(l-1:)})+\mu\dist(\hbb_{(:l+1)}^{(t)},\bb_{(:l+1)})
\ees
The last inequality holds due to the condition $\dist(\hbb_{(l-1:)}^{(t+1)},\bb_{(l-1:)}) \le \mu$ and $\dist(\hbb_{(:l+1)}^{(t)},\bb_{(:l+1)}) \le \mu$.
For the term A2, according to Lemma \ref{lm-delta}, we have
\bes &&\left\|\left(\hbSigma_{(l)}^{(t)}\right)^{-1}\left(\langle\hbb^{(t+1)}_{(l-1:)},\bb_{(l-1:)} \rangle \left[\langle\hbb^{(t)}_{(:l+1)},\bb_{(:l+1)}\rangle\hbSigma_{(l)}^{(t)}-\bSigma_{(l)}^{(t)}\right]\right)\right\|_2 \cr
&=& \langle\hbb^{(t+1)}_{(l-1:)},\bb_{(l-1:)} \rangle\left\|\left(\hbSigma_{(l)}^{(t)}\right)^{-1}\left( \langle\hbb^{(t)}_{(:l+1)},\bb_{(:l+1)} \rangle\hbSigma_{(l)}^{(t)}-\bSigma_{(l)}^{(t)}\right)\right\|_2 \cr
&\le&\frac{3\delta}{1-3\delta}\langle\hbb^{(t+1)}_{(l-1:)},\bb_{(l-1:)} \rangle \ \dist\left(\hbb_{(:l+1)}^{(t)},\bb_{(:l+1)}\right) \cr&\le&\frac{3\delta}{1-3\delta}\dist\left(\hbb_{(:l+1)}^{(t)},\bb_{(:l+1)}\right). 
\ees
For the term A3, we similarly have
\bes
\left\|\left(\hbSigma_{(l)}^{(t)}\right)^{-1}\left(\langle \hbb^{(t+1)}_{(l-1:)},\hbb_{(l-1:)} \rangle\bSigma_{(l)}^{(t)}-\bTheta_{(l)}^{(t)}\right)\right\|_2 \le\frac{3\delta}{1-3\delta}\dist\left(\hbb_{(l-1:)}^{(t+1)},\bb_{(l-1:)}\right).
\ees
For the term A4, we first note that
$\left\|\hbSigma_{(l)}^{(t)}\right\|_2^{-1}\le (1-3\delta)^{-1}$. Moreover,
\bes
&& \left\|\frac{1}{n}\hbX_{(l)}^\T\bep\right\|_2 =\left\|\frac{1}{n}\sum_{i=1}^n\eps_i\tbX_i\left(\hbb_{(:l+1)}^{(t)},  \hbb_{(l-1:)}^{(t+1)}\right) \right\|_2 \cr &\le&\sup\left\{\frac{1}{n^2}\left\|\sum_{i=1}^n\eps_i\tbX_i\left(\bb_{(:l+1)},  \bb_{(l-1:)}\right) \right\|_2,  \ \|\bb_{(:l+1)}\|_2=\|\bb_{(l-1:)}\|_2=1\right\}=\tau_0.
\ees
As a result, A4 could be bounded by
\bes
(1/\lambda)\left\|\left(\hbSigma_{(l)}^{(t)}\right)^{-1}\hbX_{(l)}^\T\bE\right\|_2\le \frac{\tau_0}{\lambda(1-3\delta)}=\tau.
\ees
Combining A1 to A4, we have
\begin{align}\label{pf-main-1}
(1/\lambda)\|\hlambda^{(t+1)}\hbb^{(t+1)}_l-\lambda\bb_l\|_2 
\le\nu \ \left[\dist(\hbb_{(l-1:)}^{(t+1)},\bb_{(l-1:)}) + \dist(\hbb_{(:l+1)}^{(t)},\bb_{(:l+1)})\right]+\tau,
\end{align}
where we recall that $\nu=\mu + 3\delta/(1-3\delta) $. 
On the other hand, due to Lemma \ref{lm-0},
\begin{align}\label{pf-main-2}
\dist(\hbb_l^{(t+1)},\bb_l)
=\dist(\hlambda^{(t+1)}\hbb_l^{(t+1)}, \ \lambda\bb_l) 
\le\frac{\|\hlambda^{(t+1)}\hbb_l^{(t+1)}-\lambda\bb_l\|_2}{\lambda}.
\end{align}
It then follows from (\ref{pf-main-1}) and (\ref{pf-main-2}) that
\begin{align}\label{pf-main-3}
\dist(\hbb_l^{(t+1)},\bb_l)\le\nu\left(\dist(\hbb_{(l-1:)}^{(t+1)},\bb_{(l-1:)}) + \dist(\hbb_{(:l+1)}^{(t)},\bb_{(:l+1)})\right)+\tau.
\end{align}
This is the central inequality. Note that for the special case with $l=1$ and $l=L$, the central inequality reduces to
\bes
\dist(\hbb_1^{(t+1)},\bb_1)&\le&
\nu\ \dist(\hbb_{(:2)}^{(t)},\bb_{(:2)}) + \tau ,\cr
\dist(\hbb_L^{(t+1)},\bb_L)&\le&
\nu\ \dist(\hbb_{(L-1:)}^{(t+1)},\bb_{(L-1:)}) + \tau.
\ees
\hfill$\square$
\bigskip

\begin{lemma}\label{lemma:(l:)}
For any given $t\ge 0$, assume $\dist\left(\hbb_{(:l)}^{(t)}, \bb_{(:l)}\right) \le \mu$ holds for all $l = 2, \cdots, L$. Let $\nu=\mu + 3\delta/(1-3\delta)$, $\tau=(\tau_0/\|\bmC\|_F)(1-3\delta)^{-1}$ and $\eta = \mu/(\mu + \tau/\nu)$. Suppose $\nu$ satisfies $(\nu+1)^{L-1}-1 < \eta$. Suppose the RIP condition holds.
Then, for all $l=1, 2, \ldots, \ L$, we have
\bes
\dist\left(\hbb_{(l-1:)}^{(t+1)}, \bb_{(l-1:)}\right) \le \mu
\ees
\end{lemma}
\bigskip

\noindent{\bf Proof of Lemma \ref{lemma:(l:)}.} We will prove a shaper inequality, Then Lemma \ref{lemma:(l:)} follows immediately. We will show that
\begin{align}\label{dist(l:)}
\dist\left(\hbb_{(l-1:)}^{(t+1)}, \bb_{(l-1:)}\right) \le \left[(\nu+1)^{l-1}-1\right]\mu + \frac{(\nu+1)^{l-1}-1}{\nu}\tau
\end{align}

We prove by induction. When $l=1$, $\dist\left(\hbb_{(0:)}^{(t+1)}, \bb_{(0:)}\right) = 0 \le 0$ holds immediately. Then suppose the statement holds for $l$, we prove it holds for $l+1$.
First note that $\dist\left(\hbb_{(l-1:)}^{(t+1)}, \bb_{(l-1:)}\right) \le \mu$ because
\bes
\dist\left(\hbb_{(l-1:)}^{(t+1)}, \bb_{(l-1:)}\right) &\le& \left[(\nu+1)^{l-1} - 1\right]\mu + \frac{(\nu+1)^{l-1}-1}{\nu}\tau \cr
&\le& \left[(\nu+1)^{L-1} - 1\right]\left(\mu + \frac{\tau}{\nu}\right) \cr
&\le& \mu.
\ees
Combining the assumption $\dist\left(\hbb_{(:l+1)}^{(t)}, \bb_{(:l+1)}\right) \le \mu$, we have inequality (\ref{central_l}) in Lemma \ref{t+1} holds. Furthermore,
\bes
&&\dist\left(\hbb_{(l:)}^{(t+1)}, \bb_{(l:)}\right) \cr
&\le& \dist\left(\hbb_{l}^{(t+1)}, \bb_{l}\right) + \dist\left(\hbb_{(l-1:)}^{(t+1)}, \bb_{(l-1:)}\right) \cr
&\le& (\nu+1)\dist\left(\hbb_{(l-1:)}^{(t+1)}, \bb_{(l-1:)}\right) + \nu\dist\left(\hbb_{(:l+1)}^{(t)}, \bb_{(:l+1)}\right) + \tau \cr
&\le& (\nu+1)\left[(\nu+1)^{l-1} - 1\right]\mu  + (\nu+1)\frac{(\nu+1)^{l-1}-1}{\nu}\tau + \nu\mu+ \tau\cr
&=& \left[(\nu+1)^{l} - 1\right]\mu + \frac{(\nu+1)^l-1}{\nu}\tau
\ees
These inequalities hold in turn by 1) Lemma \ref{lm-0}, 2) inequality (\ref{central_l}) and 3) induction holds for $l$. Thus, the statement holds for $l+1$. As a consequence, we complete the proof of (\ref{dist(l:)}).
Finally, $\dist\left(\hbb_{(l-1:)}^{(t+1)}, \bb_{(l-1:)}\right) \le \mu$ for all $l=1, \cdots, L$.
\hfill$\square$
\bigskip
\begin{lemma}\label{lemma:(:l)}
For any given $t\ge 0$, assume $\dist\left(\hbb_{(:l)}^{(t)}, \bb_{(:l)}\right) \le \mu$ holds for all $l = 2, \cdots, L$. Let $\nu=\mu + 3\delta/(1-3\delta)$, $\tau=(\tau_0/\|\bmC\|_F)(1-3\delta)^{-1}$ and $\eta = \mu/(\mu + \tau(\nu+1)/\nu)$. Suppose $\nu$ satisfies $(\nu+1)^{L-1}-1 < \eta$. Suppose the RIP condition holds.
Then
\bes
\dist\left(\hbb_{(:l)}^{(t+1)}, \bb_{(:l)}\right) \le \mu
\ees
holds for all $l=2, \ldots, \ L$.
\end{lemma}
\bigskip

\noindent{\bf Proof of Lemma \ref{lemma:(:l)}.} Similar to Lemma \ref{lemma:(l:)}, we prove the following shaper inequality holds by  induction:
\begin{align}\label{dist(:l)}
\dist\left(\hbb_{(:l)}^{(t+1)}, \bb_{(:l)}\right)
\le \left[(\nu+1)^{L-1}-(\nu+1)^{l-1}+\nu\right]\mu + \frac{(\nu+1)^{L} - (\nu+1)^{l-1}}{\nu}\tau.
\end{align}
Before that, because conditions in Lemma \ref{lemma:(:l)} are also satisfied by Lemma \ref{lemma:(l:)}, we have the inequality (\ref{dist(l:)}) holds. Additionally, with the assumption $\dist\left(\hbb_{(:l)}^{(t)}, \bb_{(:l)}\right) \le \mu$, Lemma \ref{t+1} also holds. 


For this induction, we start with $l=L$.
\bes
\dist\left(\hbb_{(:L)}^{(t+1)}, \bb_{(:L)}\right) &=& \dist\left(\hbb_{L}^{(t+1)}, \bb_{L}\right) \cr
&\le& \nu \dist\left(\hbb_{(L-1:)}^{(t+1)}, \bb_{(L-1:)}\right) + \tau\cr
&\le& \nu\left[(\nu+1)^{L-1} - 1\right]\mu + \left[(\nu+1)^{L-1}-1\right]\tau + \tau \cr
&\le& \nu\mu + (\nu+1)^{L-1}\tau
\ees
These inequalities hold in turn by 1) Lemma \ref{t+1} , 2) inequality (\ref{dist(l:)}) and 3) $(\nu+1)^{L-1} - 1 <\eta<1$. Note that $\nu\mu = \left((\nu+1)^{L-1}-(\nu+1)^{L-1}+\nu\right)\mu$ and $(\nu+1)^{L-1}\tau = \left((\nu+1)^{L} - (\nu+1)^{L-1}\right)/\nu\cdot\tau$. Thus the statement holds for $l=L$.

Next we suppose the statement holds for $l$, to prove it holds for $l-1$.
\bes
&& \dist\left(\hbb_{(:l-1)}^{(t+1)}, \bb_{(:l-1)}\right) \cr
&\le& \dist\left(\hbb_{(:l)}^{(t+1)}, \bb_{(:l)}\right) + \dist\left(\hbb_{l-1}^{(t+1)}, \bb_{l-1}\right) \cr
&\le& \dist\left(\hbb_{(:l)}^{(t+1)}, \bb_{(:l)}\right) + \nu\left(\dist\left(\hbb_{(l-2:)}^{(t+1)}, \bb_{(l-2:)}\right) + \dist\left(\hbb_{(:l)}^{(t)}, \bb_{(:l)}\right) \right) + \tau\cr
&\le& \left[(\nu+1)^{L-1}-(\nu+1)^{l-1}+\nu\right]\mu + \frac{(\nu+1)^{L} - (\nu+1)^{l-1}}{\nu} \tau \cr
&&+ \nu\left\{\left[(\nu+1)^{l-2} - 1\right]\mu + \frac{(\nu+1)^{l-2}-1}{\nu}\tau + \mu\right\} + \tau\cr
&=& \left((\nu+1)^{L-1}-(\nu+1)^{l-2}+\nu\right)\mu + \frac{(\nu+1)^{L} - (\nu+1)^{l-2}}{\nu}\tau
\ees
These inequalities hold in turn by 1) Lemma \ref{lm-0}, 2) lemma \ref{t+1} and 3) induction in $l$ and inequality (\ref{dist(l:)}).
Thus, the inequality holds for $l-1$.
So (\ref{dist(:l)}) holds for all $l=2, \cdots, L$. Finally
\bes
&&\dist\left(\hbb_{(:l)}^{(t+1)}, \bb_{(:l)}\right) \cr
&\le& \left((\nu+1)^{L-1}-(\nu+1)^{l-1}+\nu\right)\mu + \frac{(\nu+1)^{L} - (\nu+1)^{l-1}}{\nu}\tau \cr
&=& \left((\nu+1)^{L-1}-1\right)\left(\mu + \frac{\nu+1}{\nu}\tau\right) \cr
&\le& \mu
\ees
We complete the proof of Lemma \ref{lemma:(:l)}. \hfill$\square$
\bigskip



\begin{lemma}\label{lemmatau}
Suppose the model (\ref{model00}) and RIP condition hold. Assume the noise $\eps_i$ is sub-Gaussian. Let 
\bes
\tau_0=\sup\left\{\frac{1}{n}\Big\|\sum_{i=1}^n\eps_i\tbX_i\left(\bb_{(:l+1)}, \ \bb_{(l-1:)}\right)\Big\|_2, \ \|\bb_{(:l+1)}\|_2= \|\bb_{(l-1:)}\|_2=1,\  l=1,\ldots, L \right\}.
\ees
and $\tau=(\tau_0/\|\bmC\|_F)(1-3\delta)^{-1}$. Then,
\bes
\tau=\mathcal{O}_p\left(\sqrt{\frac{\log (n)}{n}}\right).
\ees
\end{lemma}
\bigskip

\noindent{\bf Proof of Lemma \ref{lemmatau}.} By a Hoeffding-type inequality, e.g., Proposition 5.10 in \citet{vershynin2010introduction}, we have
\bes
\P\left\{\frac{1}{n^2}\left\|\sum_{i=1}^n\eps_i\tbX_i\left(\bb_{(:l+1)},  \bb_{(l-1:)}\right) \right\|_2^2\ge c_0\frac{\log n}{n} \left(\frac{1}{n}\sum_{i=1}^n\left\| \tbX_i\left(\bb_{(:l+1)},  \bb_{(l-1:)}\right)\right\|_2^2\right)\right\}\le \frac{1}{n}
\ees
holds for certain constant $c_0$. Note that we may take sup on both side of inequality inside $\P\{\}$. On the other hand,  for any $\tbX_i$,  $(1/n)\sum_{i=1}^n\left\| \tbX_i\left(\bb_{(:l+1)},  \bb_{(l-1:)}\right)\right\|_2^2$ is upper bounded due to the RIP condition. Therefore
\bes
\tau_0=\left\|\hbX_{(l)}^\T\bE\right\|_2\le c\sqrt{\frac{\log n}{n}}
\ees
holds with large probability, where $c$ is certain constant. Further we have the same order probabilistic upper bound for $\tau$.
\hfill$\square$
\bigskip
\subsection{Proof of theorems}\label{sec:ths}
\noindent{\bf Proof of Theorem \ref{lemmat}}

First, by Lemma \ref{lemma:(:l)}, when $\dist\left(\hbb_{(:l)}^{(0)}, \bb_{(:l)}\right) \le \mu$ for $l = 2, \ldots, L$, we have  $\dist\left(\hbb_{(:l)}^{(t)}, \bb_{(:l)}\right) \le \mu$ holds for all $t = 0, 1, \ldots$ and $l = 2, \ldots, L$ by a simple induction.

Next, when  $\dist\left(\hbb_{(:l)}^{(t)}, \bb_{(:l)}\right) \le \mu$ holds for all $t = 0, 1, \ldots$ and $l = 2, \ldots, L$, by Lemma \ref{lemma:(l:)}, we have 
$\dist\left(\hbb_{(l-1:)}^{(t+1)}, \bb_{(l-1:)}\right) \le \mu$ holds for all $t = 0, 1, \ldots$ and $l = 2, \ldots, L$. 

Finally when $\dist\left(\hbb_{(:l)}^{(t)}, \bb_{(:l)}\right) \le \mu$ and $\dist\left(\hbb_{(l-1:)}^{(t+1)}, \bb_{(l-1:)}\right) \le \mu$ holds for all $t = 0, 1, \ldots$ and $l = 2, \ldots, L$, we have Theorem \ref{lemmat} holds by Lemma \ref{t+1}.
\hfill$\square$
\bigskip






\noindent{\bf Proof of Theorem \ref{th-main}}

According to Theorem \ref{lemmat}, inequality (\ref{central}) holds for $t=0, 1, \ldots$ and $l=1, \ldots, L$.
\begin{eqnarray}
\dist\left(\hbb_l^{(t+1)}, \bb_l\right)&\le& \nu\left(\dist\left(\hbb_{(l-1:)}^{(t+1)}, \bb_{(l-1:)}\right) + \dist\left(\hbb_{(:l+1)}^{(t)}, \bb_{(:l+1)}\right)\right)+\tau \cr
&\le&\nu\sum_{j=1}^{l-1}\dist\left(\hbb_{j}^{(t+1)}, \bb_{j}\right) + \nu\sum_{j=l+1}^{L}\dist\left(\hbb_{j}^{(t)}, \bb_{j}\right) + \tau \label{central_sum}
\end{eqnarray}
The second inequality holds by Lemma \ref{lm-0}.
Now we show that the following inequality follows from (\ref{central_sum}):
\begin{eqnarray}\label{s11}
\dist\left(\hbb_{l}^{(t+1)}, \bb_l\right) \le \sum_{j=1}^L f(l, j) \dist\left(\hbb_j^{(t)}, \bb_j\right) + h(l)\tau
\end{eqnarray}
where $f(l, j)$ and $h(l)$ are the coefficients function with the following form
\begin{eqnarray}
&&f(l, j) = \nu\left((\nu+1)^{l-1} - I[l\ge j](\nu+1)^{l-j}\right)
\quad \text{for}\ l, \ j = 1, \cdots, L \label{coef_f} \\
&&h(l) = (\nu+1)^{l-1} 
\quad \text{for}\ l = 1, \cdots, L \label{coef_h}
\end{eqnarray}
Again we prove (\ref{s11}) by induction. For $l=1$,
\bes
\dist\left(\hbb_1^{(t+1)}, \bb_1\right)&\le& \nu\dist\left(\hbb_{(:2)}^{(t)}, \bb_{(:2)}\right)+\tau \le \sum_{j=2}^L \nu\dist\left(\hbb_{j}^{(t)}, \bb_{j}\right)+\tau
\ees
which indicates that $f(1, 1) = 0$ and $f(1, j) = \nu$ for $j = 2, \cdots, L$, satisfying formulation (\ref{coef_f}). While $h(1) = 1$ also satisfies formulation (\ref{coef_h}).

Now suppose inequality (\ref{s11}) holds for $l$, we prove it holds for $l+1$.
\bes
&&\dist\left(\hbb_{l+1}^{(t+1)}, \bb_{l+1}\right) \cr
&\le& \nu\sum_{j=1}^{l}\dist\left(\hbb_{j}^{(t+1)}, \bb_{j}\right) + \nu\sum_{j=l+2}^{L}\dist\left(\hbb_{j}^{(t)}, \bb_{j}\right) + \tau \cr
&\le& \nu\sum_{j=1}^{l}\left(\sum_{k=1}^L f(j, k) \dist\left(\hbb_k^{(t)}, \bb_k\right) + h(j)\tau\right) + \nu\sum_{j=l+2}^{L}\dist\left(\hbb_{j}^{(t)}, \bb_{j}\right) + \tau \cr
&=& \nu\sum_{k=1}^{L}\sum_{j=1}^l f(j, k) \dist\left(\hbb_k^{(t)}, \bb_k\right) + \nu\sum_{j=l+2}^{L}\dist\left(\hbb_{j}^{(t)}, \bb_{j}\right) + \left(\nu\sum_{j=1}^{l}h(j)+1\right)\tau \cr
&=& \sum_{k=1}^{L}\nu\left(\sum_{j=1}^l f(j, k) + I[k\ge l+2] \right)\dist\left(\hbb_k^{(t)}, \bb_k\right) + \left(\nu\sum_{j=1}^{l}h(j)+1\right)\tau \cr
&=& \sum_{j=1}^{L}\nu\left(\sum_{k=1}^l f(k, j) + I[j\ge l+2] \right)\dist\left(\hbb_j^{(t)}, \bb_j\right) + \left(\nu\sum_{j=1}^{l}h(j)+1\right)\tau
\ees
The first two inequality holds due to 1) inequality (\ref{central_sum}) and 2) induction for $l$ respectively.

Now we compare the coefficients for $h(\cdot)$ and $f(\cdot,\cdot)$. First, the coefficient of $\tau$ can be written as $h(l+1)$ because
\bes
\nu\sum_{j=1}^{l}h(j)+1 = \nu\sum_{j=1}^{l}(\nu+1)^{j-1}+1 = (\nu+1)^l = h(l+1)
\ees

For the coefficient of $\dist\left(\hbb_j^{(t)}, \bb_j\right)$, we consider different situations. When $j \le l$:
\bes
&&\nu\left(\sum_{k=1}^l f(k, j) + I[j\ge l+2] \right) \cr
&=& \nu\left(\sum_{k=1}^{l}\nu\left((\nu+1)^{k-1} - I[k\ge j](\nu+1)^{k-j}\right)\right)\cr
&=&\nu^2\left(\sum_{k=1}^{l}(\nu+1)^{k-1} - \sum_{k=j}^{l}(\nu+1)^{k-j}\right) \cr
&=&\nu^2\sum_{k=l-j+1}^{l-1}(\nu+1)^{k}\cr
&=&\nu\left((\nu+1)^l - (\nu+1)^{l-j+1}\right)
\ees
\noindent When $j = l+1$:
\bes
\nu\left(\sum_{k=1}^l f(k, j) + I[j\ge l+2] \right)=\nu\sum_{k=1}^{l}\nu(\nu+1)^{k-1}=\nu\left((\nu+1)^l-1\right)
\ees
\noindent When $j \ge l+2$:
\bes
\nu\left(\sum_{k=1}^l f(k, j) + I[j\ge l+2] \right)=\nu\left(\sum_{k=1}^{l}\nu(\nu+1)^{k-1} + 1\right)=\nu(\nu+1)^l
\ees
In summary,
\bes
\nu\left(\sum_{j=1}^l f(j, k) + I[k\ge l+2] \right) &=& \nu\left((\nu+1)^{l} - I[l+1\ge j](\nu+1)^{l+1-j}\right) \cr
&=& f(l+1, j)
\ees
So we have 
\bes
\dist\left(\hbb_{l+1}^{(t+1)}, \bb_{l+1}\right) \le \sum_{j=1}^L f(l+1, j) \dist\left(\hbb_j^{(t)}, \bb_j\right) + h(l+1)\tau
\ees
This above proves (\ref{s11}). Specially, note that
\begin{itemize}
\item $f(l, 1) = 0$ for $l = 1, \cdots, L$.
\item $f(1, j) = \nu$ for $j = 2, \cdots, L$.
\item Define the summation and get the form
\bes
f(:, j) &=& \sum_{l=1}^L f(l, j) = \nu\left(\sum_{l=1}^L(\nu+1)^{l-1} - \sum_{l=j}^L(\nu+1)^{L-j}\right)\cr
&=& (\nu+1)^L - (\nu+1)^{L-j+1}
\ees
which is increasing with regard to $j$, so $f(:, j) \le f(:, L) = (\nu+1)^L - (\nu+1)$.
\end{itemize}


Denote $\kappa = (\nu+1)^L - (\nu+1) - \nu$. By summarizing inequalities (\ref{s11}) for $l = 2, \cdots, L$, we can get:
\begin{eqnarray}
\sum_{l=2}^{L}\dist\left(\hbb_l^{(t+1)},\bb_l\right)&\le&\sum_{l=2}^{L}\sum_{j=2}^{L}f(l, j)\dist\left(\hbb_{j}^{(t)}, \bb_j\right) + \sum_{l=2}^{L}h(l)\tau \cr
&=& \sum_{j=2}^{L}\left(f(:, j)-f(1, j)\right)\dist\left(\hbb_{j}^{(t)}, \bb_j\right) + \frac{\nu+\kappa}{\nu}\tau \cr
&\le& \kappa\sum_{l=2}^{L}\dist\left(\hbb_{l}^{(t)}, \bb_l\right) + \frac{\nu+1}{\nu}\tau \label{expansion_2}
\end{eqnarray}
These rows hold in turn as 1) inequality (\ref{s11}), 2) swap for summation order, definition of $f(:, j)$ and summation of proportional series, 3) $f(:, j) \le (\nu+1)^L - (\nu+1)$, $f(1, j) = \nu$ and $(\nu+1)^L - (\nu+1) = \nu+\kappa < \nu+1$.

Further, apply inequality (\ref{expansion_2}) $t$ times, we have
\begin{eqnarray}
\sum_{l=2}^{L}\dist\left(\hbb_l^{(t+1)},\bb_l\right)&\le& \kappa^{t+1}\sum_{l=2}^{L}\dist\left(\hbb_{l}^{(0)}, \bb_l\right) + \sum_{s=1}^{t+1}\kappa^{s-1}\frac{\nu+1}{\nu}\tau \cr
&\le& \kappa^{t+1}(L-1)\mu + \frac{\nu+1}{\nu(1-\kappa)}\tau \label{recursion}
\end{eqnarray}
These inequalities hold in turn as 1) inequality (\ref{expansion_2}) and 2) $\dist\left(\hbb_{l}^{(0)}, \bb_l\right) \le \max\left\{\dist\left(\hbb_{l}^{(0)}, \bb_l\right)\right\}_{l=2}^L \le \dist\left(\hbb_{(:2)}^{(0)}, \bb_{(:2)}\right) \le \mu$ for $l = 2, \cdots, L$.

For the first term $\dist\left(\hbb_1^{(t+1)}, \bb_1\right)$, it holds that
\bes
\dist\left(\hbb_1^{(t+1)}, \bb_1\right) \le \nu\sum_{l=2}^{L}\dist\left(\hbb_l^{(t)},\bb_l\right) + \tau \le \nu\left(\kappa^{t}(L-1)\mu + \frac{\nu+1}{\nu(1-\kappa)}\tau\right) + \tau
\ees
These inequalities hold as 1) inequality (\ref{central_sum}) and 2) inequality (\ref{recursion}).

Now add $\dist\left(\hbb_1^{(t+1)}, \bb_1\right)$ to (\ref{recursion}), we have
\bes
\sum_{l=1}^{L}\dist\left(\hbb_l^{(t+1)},\bb_l\right) &\le& \kappa^{t+1}\left(1 + \frac{\nu}{\kappa}\right)(L-1)\mu+\left(\frac{(\nu+1)^2}{\nu(1-\kappa)} + 1\right)\tau \cr
&=& c_1\kappa^{t+1}\mu + c_2\tau 
\ees
where $c_1 = (L-1)\left(1 + \frac{\nu}{\kappa}\right)$ and $ c_2 = \frac{(\nu+1)^2}{\nu(1-\kappa)} + 1$. On the other hand,
\bes
\dist\left(\widehat{\bmC}^{(t+1)}, \bmC\right) = \dist\left(\hbb_{(L:)}^{(t+1)},\bb_{(L:)}\right) \le \sum_{l=1}^{L}\dist\left(\hbb_l^{(t+1)},\bb_l\right)
\ees
\hfill$\square$

\noindent{\bf Proof of Theorem \ref{init}}

For initialization $\hbb_{(:l)}^{(0)}$ from equation (\ref{ini2}), denote
\bes
&&\tbX = \tbX^{(l)} = \left( \tvec\left(\tbX_1^{(l)}\right), \cdots, \tvec\left(\tbX_n^{(l)}\right)\right)^\T \in \mathbb{R}^{n\times dp}\cr
&&\tbX_i^{(l)} = \mR_{\left(d_{(:l)}, p_{(:l)}\right)}\left(\bX_i\right) \in \mathbb{R}^{d_{(:l)}p_{(:l)}\times d_{(l-1:)}p_{(l-1:)}}
\ees

On one hand, we have the following expansion
\bel{init11}
&&\left\|\tbX\tvec\left(\hbb_{(:l)}^{(0)}\left(\hbb_{(l-1:)}^{(0)}\right)^\T\right) - \by \right\|_2^2 \cr
&=& \left\|\tbX\tvec\left(\hbb_{(:l)}^{(0)}\left(\hbb_{(l-1:)}^{(0)}\right)^\T\right) - \left(\tbX\tvec\left(\bb_{(:l)}\left(\bb_{(l-1:)}\right)^\T \right) + \bep \right) \right\|_2^2 \cr
&=& \left\|\tbX\tvec\left(\hbb_{(:l)}^{(0)}\left(\hbb_{(l-1:)}^{(0)}\right)^\T - \bb_{(:l)}\left(\bb_{(l-1:)}\right)^\T\right) - \bep \right\|_2^2 \cr
&=& \left\|\tbX\tvec\left(\hbb_{(:l)}^{(0)}\left(\hbb_{(l-1:)}^{(0)}\right)^\T - \bb_{(:l)}\left(\bb_{(l-1:)}\right)^\T\right)\right\|_2^2 \cr
&& -2\bep^\T\left(\tbX\tvec\left(\hbb_{(:l)}^{(0)}\left(\hbb_{(l-1:)}^{(0)}\right)^\T - \bb_{(:l)}\bb_{(l-1:)}^\T\right)\right) + \|\bep\|_2^2
\eel
On the other hand by lemma 2.1 of \citet{jain2010guaranteed}, it holds that:
\bel{init12}
\left\|\tbX\tvec\left(\hbb_{(:l)}^{(0)}\left(\hbb_{(l-1:)}^{(0)}\right)^\T\right) - \by \right\|_2^2 \le
\|\bep\|_2^2 + \frac{\delta}{1-\delta}\left\|\tbX\tvec\left(\bb_{(:l)}\bb_{(l-1:)}^\T\right) \right\|_2^2
\eel
By the equality in (\ref{init11}) and inequality in (\ref{init12}), it follows that
\bes
&&\left\|\tbX\tvec\left(\hbb_{(:l)}^{(0)}\left(\hbb_{(l-1:)}^{(0)}\right)^\T - \bb_{(:l)}\bb_{(l-1:)}^\T\right)\right\|_2^2 \cr
&\le& \frac{\delta}{1-\delta}\left\|\tbX\tvec\left(\bb_{(:l)}\bb_{(l-1:)}^\T\right) \right\|_2^2 + 2\bep^\T\left(\tbX\tvec\left(\hbb_{(:l)}^{(0)}(\hbb_{(l-1:)}^{(0)})^\T - \bb_{(:l)}\bb_{(l-1:)}^\T\right)\right)
\ees

In the meantime, by the RIP condition, we have
\begin{itemize}
\item $\left\|\tbX\tvec(\bb_{(:l)}\bb_{(l-1:)}^\T)\right\|_2^2\le(1+\delta)\|\bb_{(l-1:)}\|_2^2$
\item $(1-\delta)\left\|\hbb_{(:l)}^{(0)}(\hbb_{(l-1:)}^{(0)})^\T - \bb_{(:l)}\bb_{(l-1:)}^\T\right\|_F^2 \le\left\|\tbX\tvec\left(\hbb_{(:l)}^{(0)}(\hbb_{(l-1:)}^{(0)})^\T - \bb_{(:l)}\bb_{(l-1:)}^\T\right)\right\|_2^2
\le (1+\delta)\left\|\hbb_{(:l)}^{(0)}(\hbb_{(l-1:)}^{(0)})^\T - \bb_{(:l)}\bb_{(l-1:)}^\T\right\|_F^2 $
\end{itemize}
After replacing the terms of $\tbX$, we get the following quadratic inequality
\bes
&&\left\|\hbb_{(:l)}^{(0)}\left(\hbb_{(l-1:)}^{(0)}\right)^\T - \bb_{(:l)}\bb_{(l-1:)}^\T\right\|_F^2 \cr
&\le& 2\frac{1+\delta}{1-\delta}\|\bep\|_2\left\|\hbb_{(:l)}^{(0)}\left(\hbb_{(l-1:)}^{(0)}\right)^\T - \bb_{(:l)}\bb_{(l-1:)}^\T\right\|_F + \frac{\delta(1+\delta)}{(1-\delta)^2}\|\bb_{(l-1:)}\|_2^2
\ees
Solving it gives
\bes
&&\left\|\hbb_{(:l)}^{(0)}\left(\hbb_{(l-1:)}^{(0)}\right)^\T - \bb_{(:l)}\bb_{(l-1:)}^\T\right\|_F \cr
&\le& \frac{1}{2}\left\{2\frac{1+\delta}{1-\delta}\|\bep\|_2+\sqrt{4\frac{(1+\delta)^2}{(1-\delta)^2}\|\bep\|_2^2 + 4\frac{\delta(1+\delta)}{(1-\delta)^2}\|\bb_{(l-1:)}\|^2}\right\} \cr
&\le&\frac{2(1+\delta)(\|\bep\|_2/\|\bb_{(l-1:)}\|_2)+\sqrt{\delta(1+\delta)}}{1-\delta}\|\bb_{(l-1:)}\|_2
\ees
Further,
\bes
\left\|\hbb_{(:l)}^{(0)}\left(\hbb_{(l-1:)}^{(0)}\right)^\T - \bb_{(:l)}\bb_{(l-1:)}^\T\right\|_F^2 &\ge& 
\left\|\left(\bI - \hbb_{(:l)}^{(0)}\left(\hbb_{(:l)}^{(0)}\right)^\T\right)\bb_{(:l)}\bb_{(l-1:)}^\T\right\|_F^2 \cr
&=& \|\bb_{(l-1:)}\|_2^2\left(1-\left\langle\hbb_{(:l)}^{(0)}, \bb_{(:l)}\right\rangle^2\right) \cr
&=& \|\bb_{(l-1:)}\|_2^2\dist^2\left(\hbb_{(:l)}^{(0)}, \bb_{(:l)}\right)
\ees
Combining the above two inequalities, we have
\bes
\dist\left(\hbb_{(:l)}^{(0)}, \bb_{(:l)}\right) \le \frac{2(1+\delta)\left(\|\bep\|_2/\|\bmC\|_F\right)+\sqrt{\delta(1+\delta)}}{1-\delta}
\ees
Note that $\bb_{(:l)}$ is assumed normalized so that $\|\bb_{(l-1:)}\|_2=\|\bmC\|_F=\lambda$. 
When  $\|\bep\|_2 \le c(1-\delta)\|\bmC\|_F/2$, we have
\bes
\dist\left(\hbb_{(:l)}^{(0)}, \bb_{(:l)}\right) \le \mu_0 = c(1+\delta) + \frac{\sqrt{\delta(1+\delta)}}{1-\delta}.
\ees
\hfill$\square$

\end{document}